\documentclass[10pt,twocolumn,letterpaper]{article}

\usepackage{cvpr}
\usepackage{times}
\usepackage{epsfig}
\usepackage{graphicx}
\usepackage{amsmath}
\usepackage{amssymb}
\usepackage{cuted}
\usepackage{capt-of}
\usepackage{float}
\usepackage{multirow}
\usepackage{booktabs}

\usepackage[pagebackref=true,breaklinks=true,letterpaper=true,colorlinks,bookmarks=false]{hyperref}
\cvprfinalcopy 

\def\cvprPaperID{507} 

\ifcvprfinal\fi
\begin{document}

\title{Deep Plastic Surgery: Robust and Controllable Image Editing with \\ Human-Drawn Sketches\vspace{-2mm}}

\author{Shuai Yang$^{1,2}$,~Zhangyang Wang$^2$,~Jiaying Liu$^1$~and Zongming Guo$^1$\\
$^1$ Wangxuan Institute of Computer Technology, Peking University\\
$^2$ Texas A\&M University}

\maketitle

\begin{strip}\centering
\includegraphics[width=\textwidth]{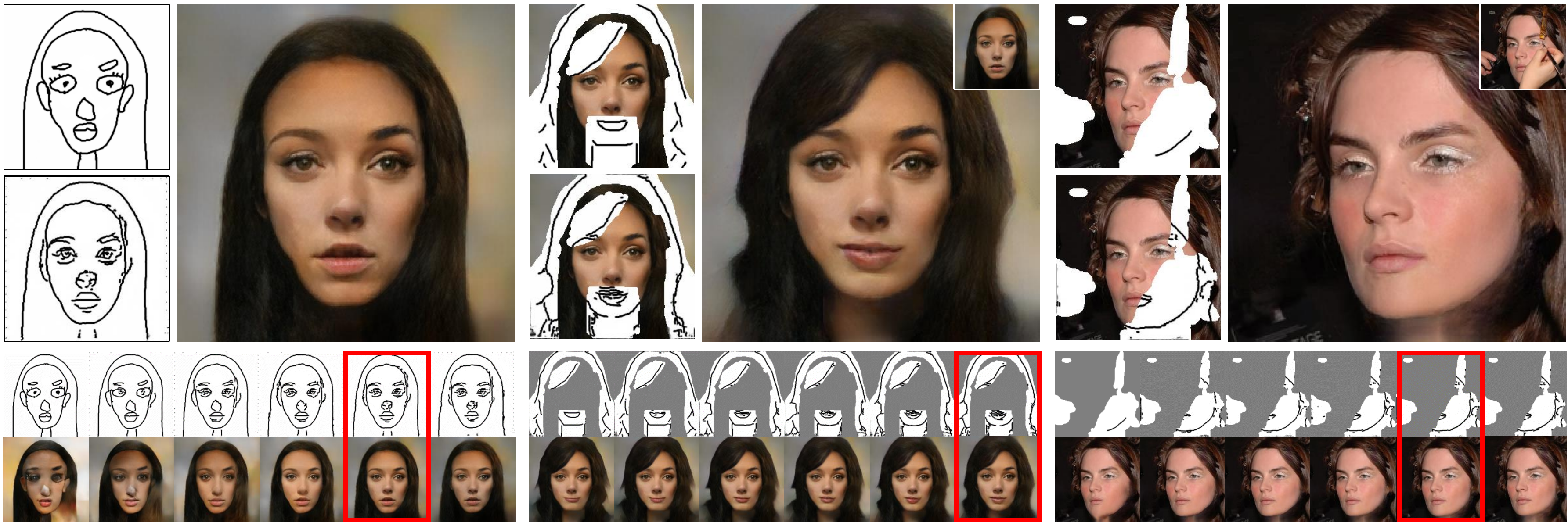}
\captionof{figure}{Our Deep Plastic Surgery framework allows users to synthesize (left) and edit (middle and right) photos based on hand-drawn sketches. Our model is robust to tolerate the drawing errors and achieves the controllability on sketch faithfulness. For each group, we show the user input and our refined sketch in the left column, and the final output in the right column with the original photo in the upper right corner. The bottom row shows our results under an increasing sketch refinement level, with a red box to indicate the user selection.}\label{fig:teaser}
\end{strip}


\begin{abstract}
  Sketch-based image editing aims to synthesize and modify photos based on the structural information provided by the human-drawn sketches. Since sketches are difficult to collect, previous methods mainly use edge maps instead of sketches to train models (referred to as edge-based models). However, sketches display great structural discrepancy with edge maps, thus failing edge-based models. Moreover, sketches often demonstrate huge variety among different users, demanding even higher generalizability and robustness for the editing model to work. In this paper, we propose Deep Plastic Surgery, a novel, robust and controllable image editing framework that allows users to interactively edit images using hand-drawn sketch inputs. We present a sketch refinement strategy, as inspired by the coarse-to-fine drawing process of the artists, which we show can help our model well adapt to casual and varied sketches without the need for real sketch training data. Our model further provides a refinement level control parameter that enables users to flexibly define how ``reliable'' the input sketch should be considered for the final output, balancing between sketch faithfulness and output verisimilitude (as the two goals might contradict if the input sketch is drawn poorly). To achieve the multi-level refinement, we introduce a style-based module for level conditioning, which allows adaptive feature representations for different levels in a singe network. Extensive experimental results demonstrate the superiority of our approach in improving the visual quality and user controllablity of image editing over the state-of-the-art methods.
\end{abstract}

\section{Introduction}



Human-drawn sketches reflect people's abstract expression of objects. They are highly concise yet expressive: usually several lines can reflect the important morphological features of an object, and even imply more semantic-level information.
Meanwhile, sketches are easily editable: such an advantage is further amplified by the increasing popularity of touch-screen devices. Sketching thus becomes one of the most important ways that people illustrate their ideas and interact with devices. Motivated by the above, a series of sketch-based image synthesis and editing methods have been proposed in recent years. The common main idea underlying these methods is to train an image-to-image translation network to map a sketch to its corresponding color image. That can be extended to image completion task where an additional mask is provided to specify the area for modification. These methods enable novice users to edit the photo through simply drawing lines, rather than resorting complicated tools to process the photo itself.


Due to the difficulty of collecting pairs of sketches and color images as training data, existing works~\cite{Isola2017Image,portenier2018faceshop,jo2019sc} typically exploit edge maps (detected from color images) as ``surrogates'' for real sketches, and train their models on the paired edge-photo datasets. Despite certain success in shoe, handbag and face synthesis, edge maps look apparently different from the human draws, the latter often being more causal, varied or even wild. As a result, those methods often generalize poorly when their inputs become human-drawn sketches, limiting their real-world usage. To resolve this bottleneck, researchers have studied edge pre-processing~\cite{portenier2018faceshop}, yet with limited performance improvement gained so far.
Some human-drawn sketch datasets have also been collected~\cite{sangkloy2016sketchy,yu2016sketch} to train sketch-based models~\cite{chen2018sketchygan,liu2019unpaired}. However, the collection is too laborious to extend to larger scales or to meet all data subject needs.


As a compromise, it is valuable to study the adaption of edge-based models to the sketches.
ContextualGAN~\cite{lu2018image} presents an intuitive solution. It retrieves the nearest neighbor of the input sketch from the learned generative edge-image manifolds, which relaxes the sketch constraint to trade for the image naturalness. However, neither edge-based models~\cite{Isola2017Image,portenier2018faceshop,jo2019sc} nor ContextualGAN~\cite{lu2018image} allows for any user controllability on the \textit{sketch faithfulness}, \ie, to what extent we should stick to the given sketch? The former categories of methods completely hinge on the input sketch even it might yield highly unnatural outputs; while the latter mainly searches from natural manifolds (with the sketch providing just a starting point) and can produce differently from the sketch specification. That leaves little room for users to calibrate between freedom of sketching and the overall image verisimilitude: an important desirable feature for interactive photo editing.




In view of the above, we are motivated to investigate a new problem of controllable sketch-based image editing, that can work robustly on varied human-drawn sketches. Our main idea is to refine the sketches to approach the structural features of the edge maps, therefore avoiding the tedious collection of sketch data for training, while enabling users to control the refinement level freely.
The challenge of this problem lies in two aspects.
\underline{First}, in the absence of real sketches as reference, we have no paired or unpaired data to directly establish a mapping between sketches and edge maps.
\underline{Second}, in order to achieve controllability, it is necessary to extend the above mapping to a multi-level progress, which remains to be an open question.

In this paper, we present \textit{Deep Plastic Surgery}, a novel sketch-based image editing framework to achieve both \textbf{robustness} on hand-drawn sketch inputs, and the \textbf{controllability} on sketch faithfulness. Our key idea arises from our observation on the coarse-to-fine drawing process of the human artists: they first draw coarse outlines to specify the approximate areas where the final fine-level lines are located. Those are then gradually refined to converge to the final sharper lines. Inspired by so, we propose a dilation-based sketch refinement method.
Instead of directly feed the network with the sketch itself, we only specify the approximate region covering the final lines, created by edge dilation, which forces the network to find the mapping between the coarse-level sketches and fine-level edges. The level of coarseness can be specified and adjusted by setting the dilation radius.
Finally, we treat sketches under different coarse levels as different stylized versions of the fine-level lines, and use the scale-aware style transfer to recover fine lines by removing their dilation-based styles.
Our method only requires color images and their edge maps to train and can adapt to diversified sketch input.
It can work as a plug-in for existing edge-based models, providing refinement for their inputs to boost their performance.

Our contributions are summarized as three-folds:\vspace{-1mm}
\begin{itemize}
  \item We explore a new problem of controllable sketch-based image editing, to adapt edge-based models to human-drawn sketches, where the users have the freedom to balance the sketch faithfulness with the output verisimilitude.\vspace{-2mm}
  \item We propose a sketch refinement method using coarse-to-fine dilations, following the drawing process of artists in real world.\vspace{-2mm}
  \item We propose a style-based network architecture, which successfully learns to refine the input sketches into diverse and continuous levels.
\end{itemize}

\section{Related Work}

\textbf{Sketch-based image synthesis}.~Generating images from sketches enables novice users to conveniently edit photos through modifying sketches. Using the easily accessible edge maps to simulate sketches, edge-based models~\cite{Isola2017Image,sangkloy2017scribbler,dekel2018sparse} are trained to map edges to their corresponding photos.
By introducing masks, they are extended to image completion tasks to modify the specified photo areas~\cite{yu2018free, portenier2018faceshop, jo2019sc} or provide users with sketch recommendations~\cite{ghosh2019isketchnfill}.
However, the drastic structural discrepancy between edges and human-drawn sketches makes the aforementioned models less generalizable to sketches. As sketches draw increasing research attentions and some datasets~\cite{sangkloy2016sketchy,yu2016sketch} are released, the discrepancy can be narrowed~\cite{chen2018sketchygan,liu2019unpaired}. But existing datasets are far from enough and collecting human-draw sketches in large scale is still too expensive.

\begin{figure*}[htbp]
\centering
\includegraphics[width=1\linewidth]{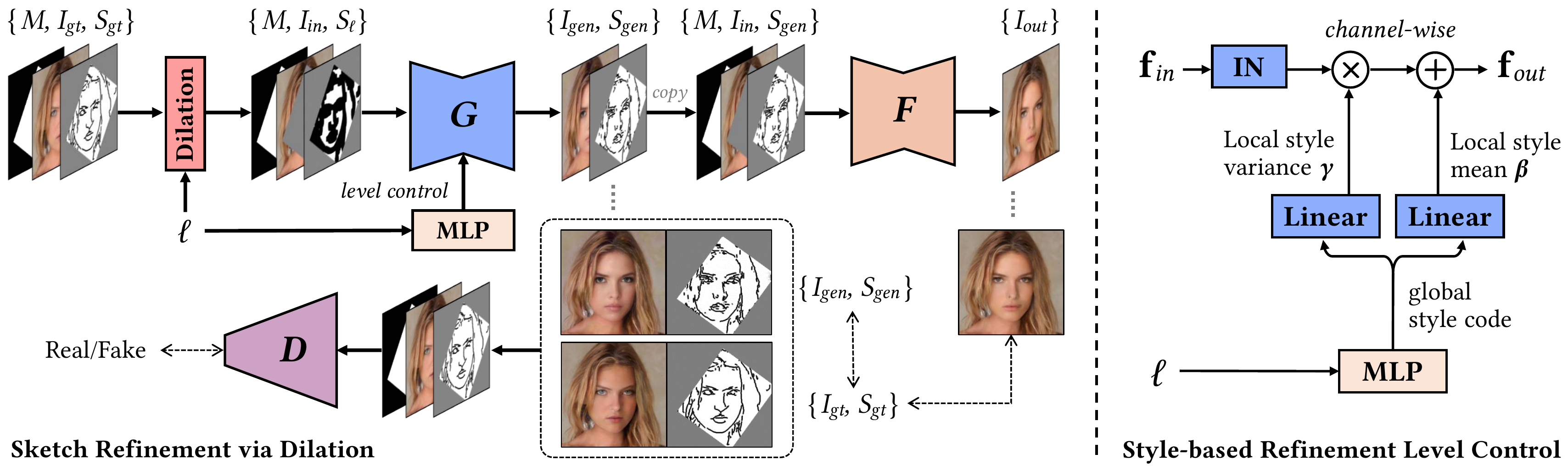}
\caption{Overview of our Deep Plastic Surgery framework. A novel sketch refinement network $G$ is proposed to refine the rough sketch $S_{\ell}$ modelled as dilated drawable regions to match the fine edges $S_{gt}$. The refined output $S_{gen}$ is fed into a pretrained edge-based model $F$ to obtain the final editing result $I_{out}$. A parameter $\ell$ is introduced to control the refinement level. It is realized by encoding $\ell$ into style codes and performing a style-based adjustment over the outputs $\mathbf{f}_{in}$ of the convolutional layers of $G$ to remove the dilation-based styles.}\vspace{-2.5mm}
\label{fig:framework}
\end{figure*}

The most related method to our problem setting is ContextualGAN~\cite{lu2018image} that also aims to adapt edge-based models to sketches.~It solves this problem by learning a generative edge-image manifold through GANs, and searching nearest neighbors to the input sketch in this manifold. As previously discussed, ContextualGAN offers no controllablity, and the influence of the sketch input might be limited for the final output. Besides, the nearest neighbor search costs time-consuming iterative back-propagation. It also relies on the generative manifolds provided by GANs, which can become hard to train as image resolution grows higher. Thus, results reported in~\cite{lu2018image} are of a limited $64\times64$ size.
By comparison, our method is able to refine $256\times256$ sketches in a fast feed-forward way, with their refinement level controllable to facilitate flexible and user-friendly image editing.

\textbf{Image-to-image translation}. Image-to-image translation networks have been proposed to translate an image from a source domain into a target domain. Isola~\textit{et al.}~\cite{Isola2017Image} designed a general image-to-image translation framework named pix2pix~\cite{Isola2017Image} to map semantic label maps or edge maps into photos. Follow-ups involve the diversification of the generated images~\cite{zhu2017toward}, high-resolution translation~\cite{Wang2017High}, and multi-domain translation~\cite{Choi2017StarGAN}.
This framework requires that images in two domains exist as pairs for training.
Zhu~\textit{et al.}~\cite{Zhu2017Unpaired} suggested a cycle consistency constraint to map the translated image back to its original version, which successfully trained CycleGAN on unpaired data. By assuming a shared latent space across two domains, UNIT~\cite{Liu2017Unsupervised} and MUNIT~\cite{huang2018multimodal} are proposed upon CycleGAN to improve the translation quality and diversity.

\textbf{Image completion}. Image completion or image inpanting aims to reconstruct the missing parts of an image. Early work~\cite{bertalmio2000image} smoothly propagates pixel values from the known region to the missing region. To deal with large missing areas, examplar-based methods are proposed to synthesize textures by sampling pixels or patches from the known region in a greedy~\cite{criminisi2004region,sun2005image} or global~\cite{wexler2007space,barnes2009patchmatch} manner.
However, the aforementioned methods only reuse information of known areas, but cannot create unseen content. In parallel, data-driven methods~\cite{hays2007scene,wang2014biggerpicture,shan2014photo} are proposed to achieve creative image completion or extrapolation by retrieving, aligning and blending images of similar scenes from external data.
Recent models such as Context Encoder~\cite{Pathak2016Context} and DeepFill~\cite{yu2018generative,yu2018free} build upon the powerful deep neural networks to leverage the extra data for semantic completion, which supports fast intelligent image editing for high-resolution images~\cite{Yang2016High} and free-form masks~\cite{yu2018free}.

\section{The Deep Plastic Surgery Algorithm}

As illustrated in Fig.~\ref{fig:framework}, given an edge-based image editing model $F$ trained on edge-image pairs $\{S_{gt}, I_{gt}\}$,
our goal is to adapt $F$ to human-draw sketches through a novel sketch refinement network $G$ that can be trained without sketch data.
$G$ aims to refine the input sketch to match the fine edge maps $S_{gt}$. The output is then fed into $F$ to obtain final editing results.
Our model is further conditioned by a control parameter $\ell\in[0,1]$ indicating the refinement level, where larger $\ell$ corresponds to greater refinement.

In Sec.~\ref{sec:dilation}, we describe our key idea of sketch refinement.
In Sec.~\ref{sec:multi-scale}, we introduce our image editing framework that enables effective controllable refinements.

\begin{figure}[t]
\centering
\includegraphics[width=1\linewidth]{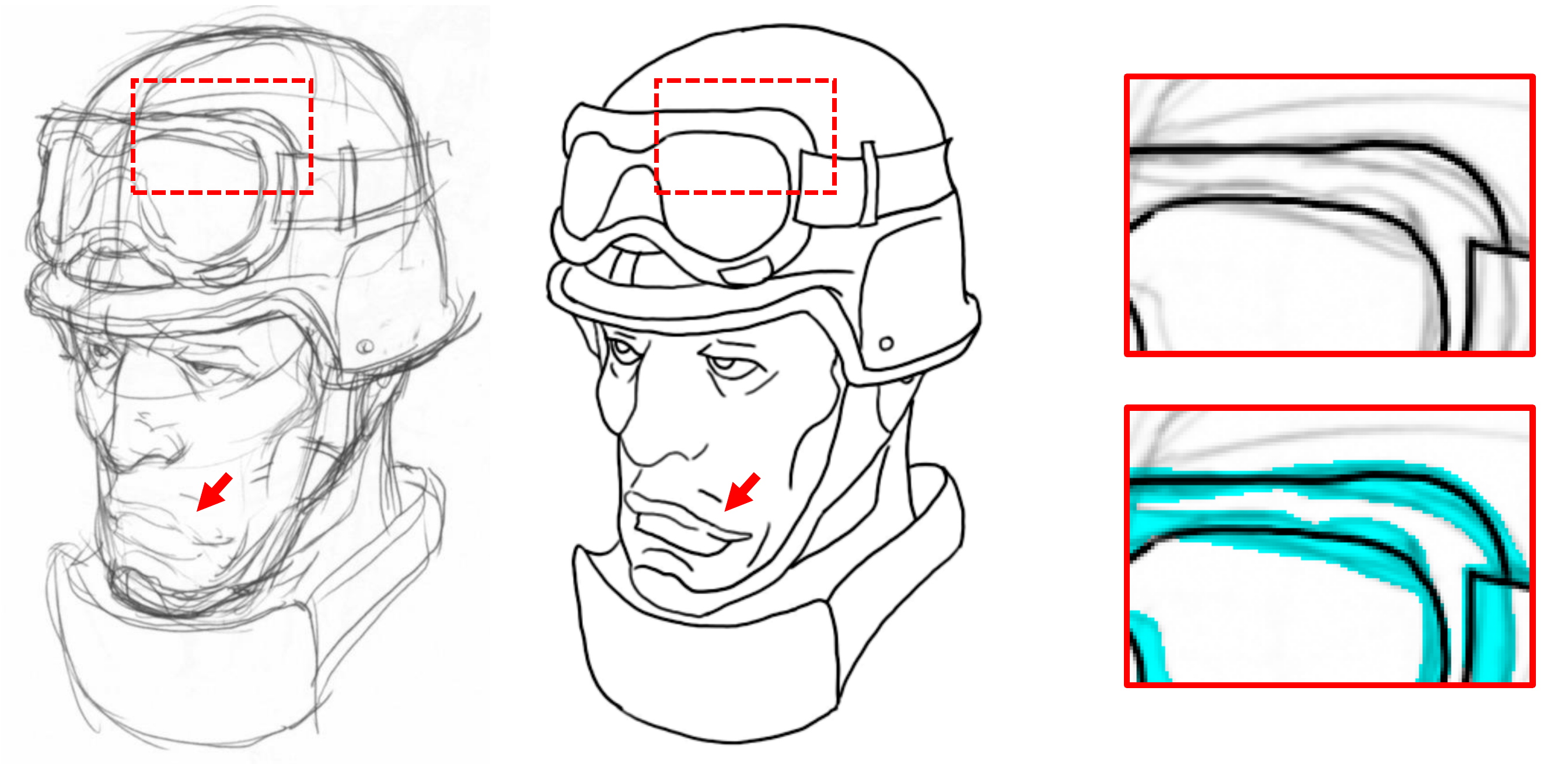}
\caption{Rough sketch (left) to fine sketch (middle). The sketches in the red boxes are enlarged and overlayed on the right. Image is from~\cite{simo2018real} and copyrighted by Krenz Cushart.}\vspace{-4mm}
\label{fig:principle}
\end{figure}

\subsection{Sketch Refinement via Dilation}
\label{sec:dilation}

Our sketch refinement method is inspired by the coarse-to-fine drawing process of human artists.
As shown in Fig.~\ref{fig:principle}, artists usually begin new illustrations with inaccurate rough sketches with many redundant lines to determine the shape of an object.~These sketches are gradually fintuned by merge lines, tweaking details and fixing mistakes to obtain the final line drawings. When overlaying the final lines on the rough sketches, we find that the redundant lines in the
rough sketches form a drawable region to indicate where the final lines should lie (tinted in cyan in Fig.~\ref{fig:principle}).~Thus \textit{the coarse-to-fine drawing process is essentially a process of continuously reducing the drawable region}.

Based on the observation, we define our sketch refinement as an image-to-image translation task between rough and fine sketches, where in our problem, fine sketches $S_{gt}$ are edge maps extracted from $I_{gt}$ using existing edge detection algorithms like HED~\cite{xie2015holistically} and \textit{rough sketches are modelled as drawable regions} $\Omega(S_{gt})$ completely covering $S_{gt}$.
In the following, we present our dilation-based drawable region generation algorithm to automatically generate $\Omega(S_{gt})$ based on $S_{gt}$ to form our training data.

\begin{figure}[t]
\centering
\includegraphics[width=1\linewidth]{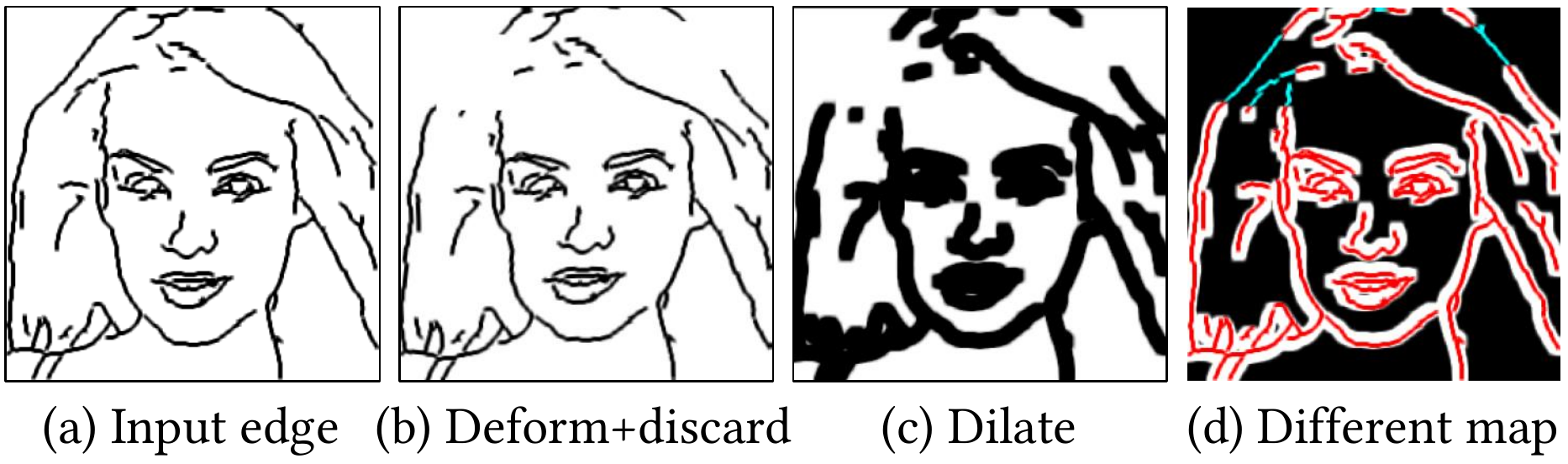}
\caption{Dilation-based rough sketch synthesis. (a) $S_{gt}$. (b) Deformed edges with lines discarded. (c) $\Omega(S_{gt})$. (d) Overlay red $S_{gt}$ above $\Omega(S_{gt})$ with discarded lines additionally tinted in cyan.}\vspace{-4mm}
\label{fig:dilate}
\end{figure}

\textbf{Rough sketch data generation}.~The pipeline of our drawable region generation is shown in Fig.~\ref{fig:dilate}.
The main idea is to expand lines into areas by dilation operations used in mathematical morphology.
However, directly learning to translate a dilated line back to itself will only make the network to simply extract the skeleton centered at the region without refining the sketches.
Thus the fine lines are first randomly deformed before dilation.
Supposing the radius of dilation is $r$, then we limit the offset of each pixel after deformation to no more than $r$, so that the ground truth fine lines are not centered at the drawable region but still fully covered by it, as shown in Fig.~\ref{fig:dilate}(d).
In addition, noticing that artists will also infer new structures or details from the draft (see the upper lip pointed by the red arrow of Fig.~\ref{fig:principle}), we further discard partial lines by removing random patches from the full sketches. By doing so, our network is motivated to learn to complete the incomplete structures such as the cyan lines in Fig.~\ref{fig:dilate}(d).~Note that the line deformation and discarding are only applied during the training phase.

\begin{figure}[t]
\centering
\includegraphics[width=1\linewidth]{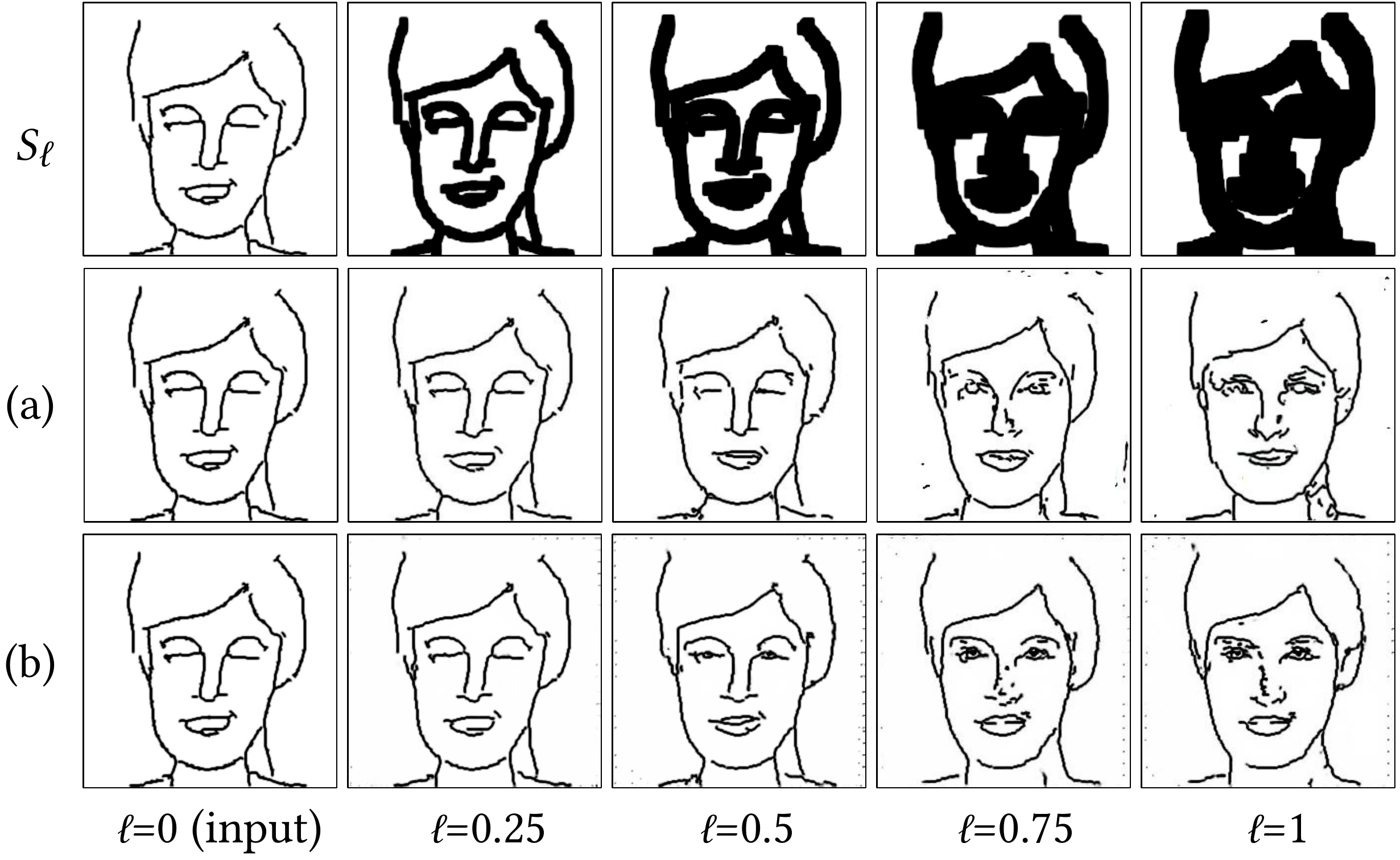}
\caption{Sketch refinement at different level $\ell$. Top row: drawable regions using different dilation radii. (a) Refinement results by pix2pix~\cite{Isola2017Image} trained separately for each level. (b) Refinement results by our proposed single model with multi-level control.}\vspace{-3mm}
\label{fig:pix2pix}
\end{figure}

Leveraging our dilation-based drawable region generation algorithm, sufficient paired data $\{\Omega(S_{gt}),S_{gt}\}$ is obtained. Intuitively, larger drawable regions provide more room for line-fintuning,~which means a higher refinement level.~To verify our idea of coarse-to-fine refinement, we train a basic image-to-image translation model of pix2pix~\cite{Isola2017Image} to map $\Omega(S_{gt})$ to $S_{gt}$ and use a separate model for each dilation radius.
As shown in Fig.~\ref{fig:pix2pix}(a), the rough facial structures are successfully refined and a growing refinement is
observed as the radius increases.
This property makes it possible for convenient sketch editing control. In the next section, we will detail how we incorporate sketch refinement into one single model with effective level control, whose overall performance is illustrated in Fig.~\ref{fig:pix2pix}(b).

\subsection{Controllable Sketch-Based Image Editing}
\label{sec:multi-scale}

In our image editing task, we have a target photo $I_{gt}$ as input, upon which a mask $M$ is given to indicate the editing region.~Users draw sketches $S$ to serve as a shape guidance for the model to fill the masked region. The model will adjust $S$ so that it better fits the contextual structures of $I_{gt}$, with the refinement level determined by a parameter $\ell$.

Our training requires no human-drawn sketches. Instead, we use edge maps $S_{gt}=\text{HED}(I_{gt})$~\cite{xie2015holistically} and generate their corresponding drawable regions $\Omega(S_{gt})$. As analyzed in Sec.~\ref{sec:dilation}, the refinement level is positively correlated with the dilation radius $r$. Therefore, we incorporate $\ell$ in the drawable region generation process (denoted as $\Omega_\ell(\cdot)$) to control $r$, where $r=\ell R$ with $R$ the maximum allowable radius.~The final drawable region with respect to $\ell$ takes the form of $S_{\ell}=\Omega_\ell(S_{gt})\odot M$ where $\odot$ is the element-wise multiplication operator.
Then we are going to train $G$ to map $S_{\ell}$ back to the fine $S_{gt}$ based on the contextual condition $I_{in}=I_{gt}\odot(\mathbf{1}-M)$, the spatial condition $M$ and the level condition $\ell$.
Fig.~\ref{fig:framework} shows an overview of our network architecture. $G$ receives a concatenation of $I_{in}$, $S_\ell$ and $M$, with middle layers controlled by $\ell$, and yields a four-channel tensor: the completed RGB channel image $I_{gen}$ and the refined one channel sketch $S_{gen}$, \textit{i.e.}, $(I_{gen},S_{gen})=G(I_{in},S_{\ell},M,\ell)$.
Here, we task the network with photo generation to enforce the perceptual guidance on the edge generation. It also enables our model to work independently if $F$ is unavailable.
Finally, a discriminator $D$ is added to improve the results through adversarial learning.

\textbf{Style-based refinement level control}.
As we will show later, conditioning by label concatenation or feature interpolation~\cite{Yang2019Controllable} fails to properly condition $G$ about the refinement level. Inspired by AdaIN-based style transfer~\cite{huang2017adain} and image generation~\cite{karras2019style}, we propose an effective style-based control module to address this issue.
Specifically, sketches at different coarse levels can be considered to have different styles. And $G$ is tasked to destylize them to obtain the original $S_{gt}$. In AdaIN~\cite{huang2017adain}, styles are modelled as the mean and variance of the features and are transferred via distribution scaling and shifting (\textit{i.e.}, normalization$+$denormalization). Note that the same operation can also be used for its reverse process, \textit{i.e.}, destylization.
To this end, as illustrated by Fig.~\ref{fig:framework}, we propose to use a multi-layer perceptron to decode the condition $\ell$ into a global style code. For each convolution layer expect the first and the last ones in $G$, we have two affiliated linear layers to map the style code to the local style mean and variance for AdaIN-based destylization. 

\textbf{Loss function}. $G$ is tasked to approach the ground truth photo and sketch in an $L_1$ sense:
\begin{equation}\label{eq:rec_loss}
\begin{aligned}
  \mathcal{L}_{\text{rec}}=\mathbb{E}_{I_{gt},M,\ell}[&~\|I_{gen}-I_{gt}\|_{1}+\|S_{gen}-S_{gt}\|_{1}\\
  +&~\|I_{out}-I_{gt}\|_{1}],
\end{aligned}
\end{equation}
where $I_{out}=F(I_{in},S_{gen},M)$ is the ultimate output in our problem. Here the quality of $I_{out}$ is also considered to adapt $G$ to the pretrained $F$ in an end-to-end manner. Besides, perceptual loss $\mathcal{L}_{\text{perc}}$~\cite{Johnson2016Perceptual}  to measure the semantical similarity of the photos is computed as
\begin{equation}\label{eq:feat_loss}
\begin{aligned}
  \mathcal{L}_{\text{perc}}=\mathbb{E}_{I_{gt},M,\ell}[\sum_i\lambda_i(&\|\Phi_i(I_{gen})-\Phi_i(I_{gt})\|^2_{2}\\
  +&\|\Phi_i(I_{out})-\Phi_i(I_{gt})\|^2_{2})],
\end{aligned}
\end{equation}
where $\Phi_i(x)$ is the feature map of $x$ in the $i$-th layer of VGG19~\cite{russakovsky2015imagenet} and $\lambda_i$ is the layer weight. Finally, we use hinge loss as our adversarial objective function:
\begin{equation}
  \mathcal{L}_{G}=-\mathbb{E}_{I_{gt},M,\ell}[D(I_{gen},S_{gen},M)],
\end{equation}
\begin{equation}\label{eq:adv_loss}
\begin{aligned}
  \mathcal{L}_{D}=&~\mathbb{E}_{I_{gt},M,\ell}[\text{ReLU}(\tau+D(I_{gen},S_{gen},M))]\\
  +&~\mathbb{E}_{I_{gt},M}[\text{ReLU}(\tau-D(I_{gt},S_{gt},M))],
\end{aligned}
\end{equation}
where $\tau$ is a margin parameter.

\textbf{Realistic sketch-to-image translation}. Under the extreme condition of $M=\textbf{1}$, $I_{gt}$ is fully masked out and our problem becomes a more challenging sketch-to-image translation problem. We experimentally find that the result will degrade without any contextual cues from $I_{gt}$.
To solve this problem, we adapt our model by removing the $I_{in}$ and $M$ inputs, and train a separate model specifically for this task, which brings obvious quality improvement.

\section{Experimental Results}

\subsection{Implementation Details}

\textbf{Dataset}. We use CelebA-HQ dataset~\cite{karras2018progressive} with edge maps extracted by HED edge detector~\cite{xie2015holistically} to train our model. All images are resized to $256\times256$ pixels. We select the first 29K images for training and the remaining 1K images for testing.
The masks are generated as the randomly rotated rectangular regions following~\cite{portenier2018faceshop}, which teaches the network to handle all possible slopes to deal with arbitrarily shaped masks during testing.
In addition, to make a fair comparison with ContextualGAN~\cite{lu2018image}, we also train our model on CelebA dataset~\cite{liu2015deep} preprocessed and provided by ContextualGAN~\cite{lu2018image}, where images are of $64\times64$ size, 196K of which are for training and 1K for testing.

\textbf{Network architecture}. Our generator $G$ utilizes the fully convolutional Encoder-ResBlocks-Decoder architecture as in~\cite{Johnson2016Perceptual}. Skip connections~\cite{Isola2017Image} are added between the Encoder and the Decoder to preserve the low-level information. Each convolutional layer is followed by AdaIN layer~\cite{huang2017adain} except the first and the last layer.
The discriminator $D$ follows the SN-PatchGAN~\cite{yu2018free} for stable and fast training. Finally, we use pix2pix~\cite{Isola2017Image} as our edge-based baseline model $F$. $F$ is trained using a standard discriminator with spectral normalization~\cite{miyato2018spectral}.
We implement dilation operations as convolutional layers with all-ones kernels of different radii $r$, followed by data clipping into the range $[0,1]$. Dilation results using the fractional radii are obtained by interpolating the results using the integer radii. More details are provided in the supplementary material.

\textbf{Network training}.
The Adam optimizer is adopted with a fixed learning rate of $0.0002$.
We first train our network with $\ell=1$ for $30$ epoches, and then train with uniformly sampled $\ell\in[0,1]$ for $200$ epoches.
The maximum allowable dilation radius is set to $R=10$ for CelebA-HQ dataset~\cite{karras2018progressive} and $R=4$ for CelebA dataset~\cite{liu2015deep}.
For all experiments, the weight for $\mathcal{L}_{\text{rec}}$, $\mathcal{L}_{\text{perc}}$, $\mathcal{L}_{G}$ and $\mathcal{L}_{\text{D}}$ are $100$, $1$, $1$ and $1$, respectively.
To calculate $\mathcal{L}_{\text{perc}}$, we use the conv2\_1 and conv3\_1 layers of the VGG19~\cite{russakovsky2015imagenet} weighted by $1$ and $0.5$, respectively.
For hinge loss, we set $\tau$ to $10$ and $1$ for $G$ and $F$, respectively.

\begin{figure*}[t]
  \centering
  \includegraphics[width=0.98\linewidth]{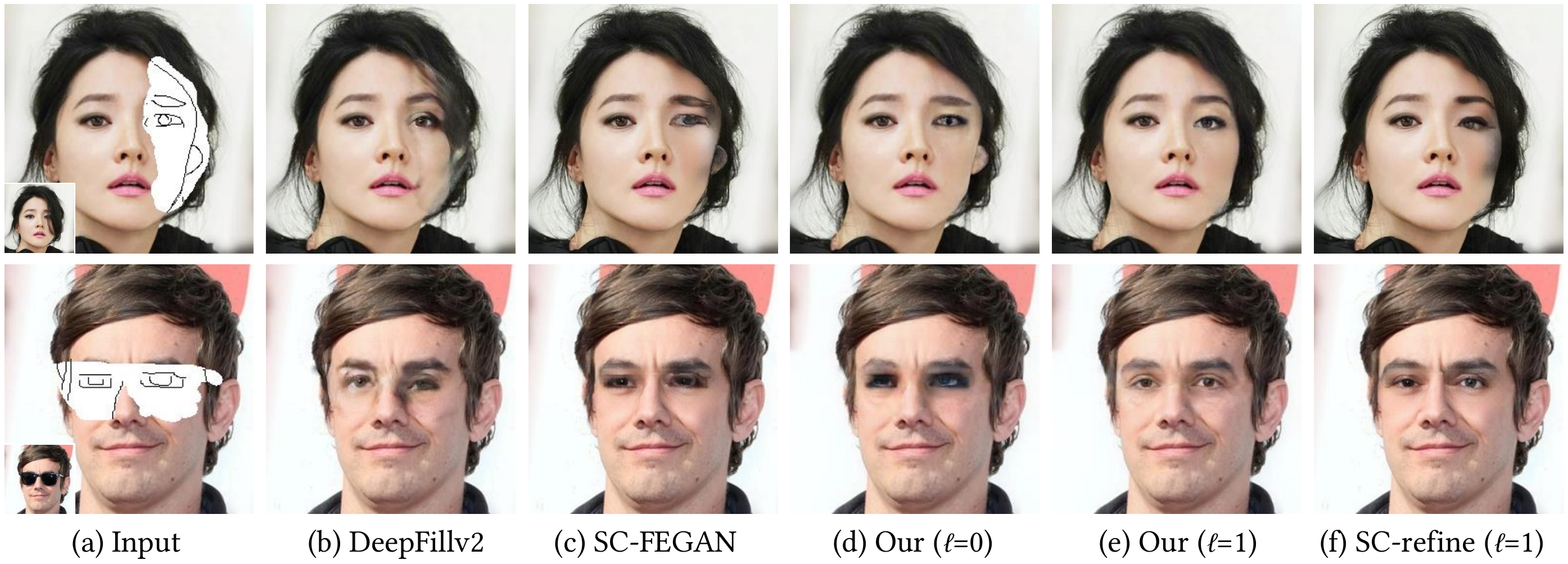}
  \caption{Comparison with state-of-the-art methods on face edting. (a) Input photos, masks and sketches. (b) DeepFillv2~\cite{yu2018free}. (c) SC-FEGAN~\cite{jo2019sc}.  (d) Our results with $\ell=0$. (e) Our results with $\ell=1$. (f) Results of SC-FEGAN using our refined sketches as input.}\label{fig:comparison-IP-1}
\end{figure*}

\begin{figure*}[t]
  \centering
  \includegraphics[width=0.98\linewidth]{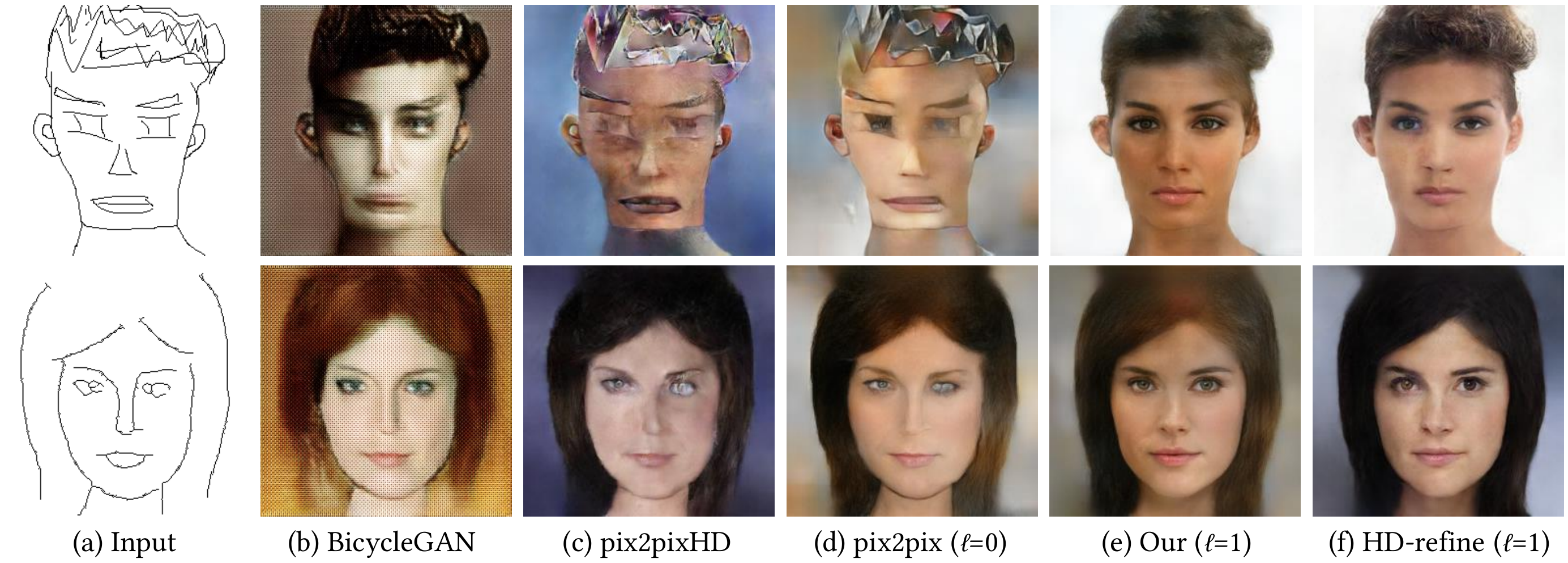}
  \caption{Comparison with state-of-the-art methods on face synthesis. (a) Input human-drawn sketches. (b) BicycleGAN~\cite{zhu2017toward}. (c) pix2pixHD~\cite{Wang2017High}.  (d) pix2pix~\cite{Isola2017Image}. (e) Our results with $\ell=1$. (f) Results of pix2pixHD using our refined sketches as input.}\vspace{-2mm}
  \label{fig:comparison-SN-1}
\end{figure*}

\subsection{Comparisons with State-of-the-Art Methods}
\vspace{-1mm}

In this section, we present comparisons with state-of-the-art methods on the tasks of face editing and synthesis. In addition to the examples shown in this paper, full results and user study to quantitatively verify the superiority of our method are included in the supplementary material.

\textbf{Face editing and synthesis}.~Fig.~\ref{fig:comparison-IP-1} presents the qualitative comparison on face editing with two state-of-the-art inpainting models: DeepFillv2~\cite{yu2018free} and SC-FEGAN~\cite{jo2019sc}. The released DeepFillv2 uses no sketch guidance, which means the reliability of the input sketch is set to zero ($\ell=\infty$). Despite being one of the most advanced inpainting models, DeepFillv2 fails to repair the fine-scale facial structures well, indicating the necessity of user guidance. SC-FEGAN, on the other hand, totally follows the inaccurate sketch and yields weird faces, due to unrealistic details contained in the rough sketches.
Similar results can be found in the output of $F$ when $\ell=0$. By using a large refinement level ($\ell=1$), the facial details become more natural and realistic.
Finally, as an ablation study to indicate the importance of sketch-edge input adaption, we directly feed SC-FEGAN with our refined sketch (without fine-tuning upon SC-FEGAN), and observe improved results of SC-FEGAN.

Fig.~\ref{fig:comparison-SN-1} shows the qualitative comparison on face synthesis with two state-of-the-art image-to-image translation models: BicycleGAN~\cite{zhu2017toward} and pix2pixHD~\cite{Wang2017High}. As expected, both models as well as $F$ (pix2pix~\cite{Isola2017Image}) synthesize facial structures that strictly match the inaccurate sketch inputs, producing poor results. Our model takes sketches as ``useful yet flexible'' constraints, and strikes a good balance between authenticity and consistency with the user guidance.

\begin{figure}[t]
  \centering
  \includegraphics[width=0.98\linewidth]{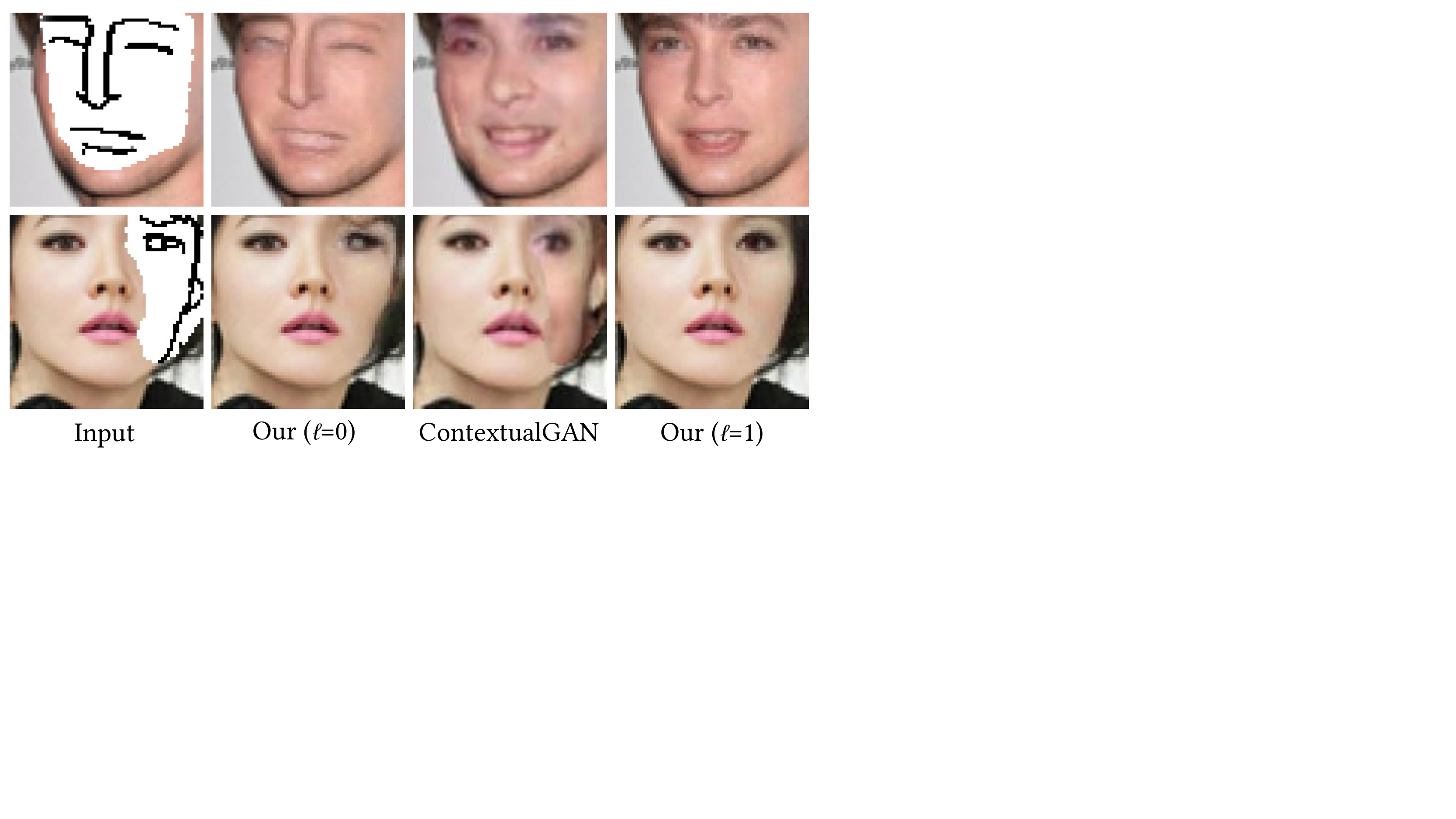}
  \caption{Comparison with ContextualGAN~\cite{lu2018image} on face editing.}\label{fig:comparison-IP-2}
\end{figure}

\begin{figure}[t]
  \centering
  \includegraphics[width=0.98\linewidth]{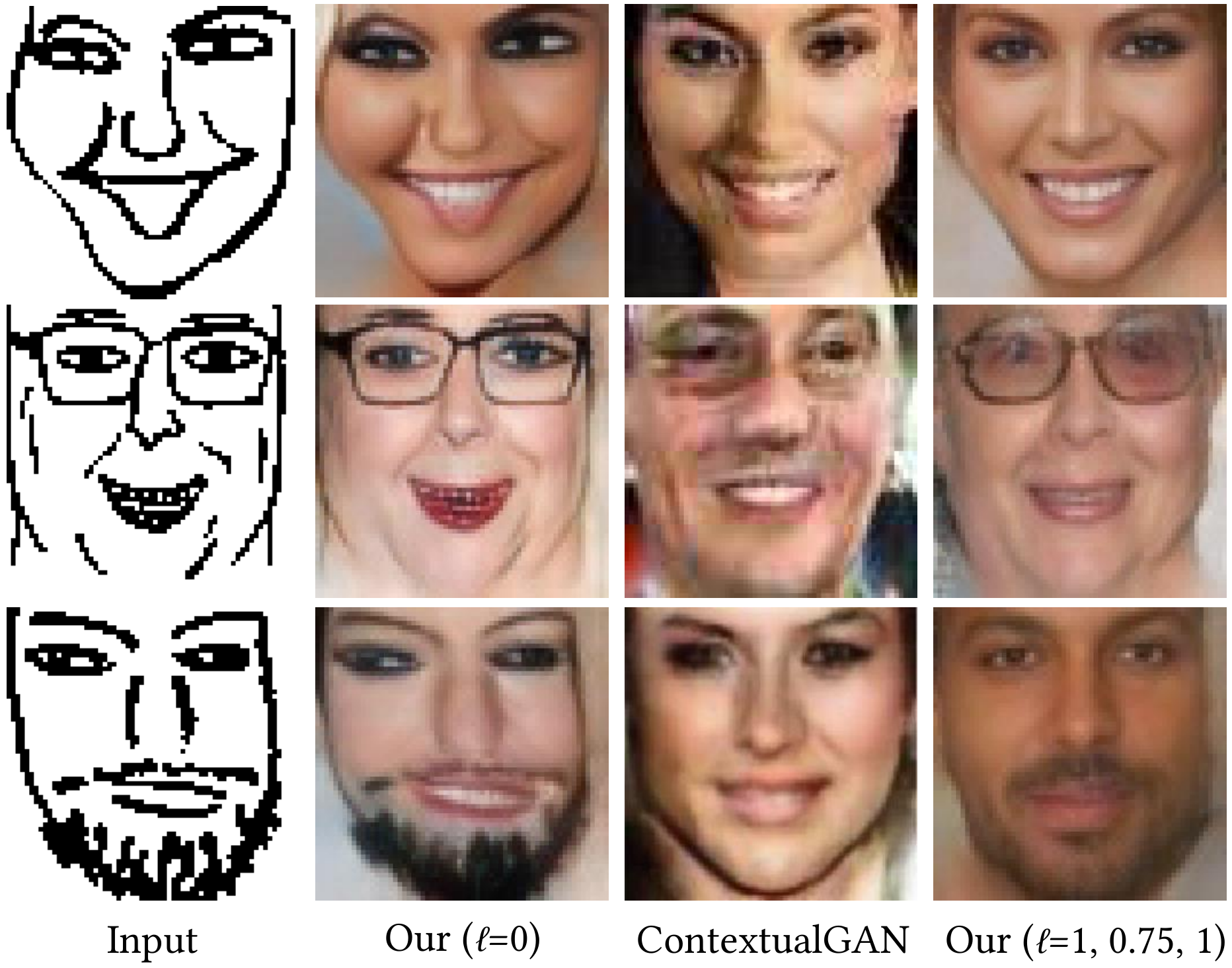}
  \caption{Comparison with ContextualGAN~\cite{lu2018image} on face synthesis. From top to bottom, we set $\ell$ to $1$, $0.75$ and $1$, respectively. }\label{fig:comparison-SN-2}
\end{figure}

\textbf{Comparison with ContextualGAN}. As the most related work that accepts weak sketch constraint as our model, we further compare with ContextualGAN~\cite{lu2018image}. For face editing task, we implement ContextualGAN and adapt it to the completion task by additionally computing the appearance similarity between the known part of the photo during the nearest neighbor search. As shown in Fig.~\ref{fig:comparison-IP-2}, the main downside of ContextualGAN is the distinct inpainting boundaries, likely due to that the learned generative manifold does not fully depict the real facial distribution.
By comparison, our method produces more natural results.

Fig.~\ref{fig:comparison-SN-2} shows
the sketch-to-image translation results, where the results of ContextualGAN are directly imported from the original paper. As can be seen, although realistic, the results of ContextualGAN lose certain attributes associated with the input such as the beard and glasses. It might be because the learned generative manifolds collapse for some uncommon attributes. As another possible cause, the nearest neighbor search might sometimes travel too far over the manifold, and results in found solutions less relevant to the initial points provided by user sketches.
Our method, by comparison, preserves these attributes much better.  

In terms of efficiency, for $64\times64$ images in Fig.~\ref{fig:comparison-SN-2}, our implemented ContextualGAN requires about 7.89 s per image with a GeForce GTX 1080 Ti GPU, while the proposed feed-forward method only takes about \textbf{12 ms per image}, which implies a potential of real-time user interaction. 

\subsection{Ablation Study}
\label{sec:ablation}

In this section, we perform ablation studies to verify our model design. We test on the challenging sketch-to-image translation task for better comparison.

\begin{figure}[t]
  \centering
  \includegraphics[width=0.98\linewidth]{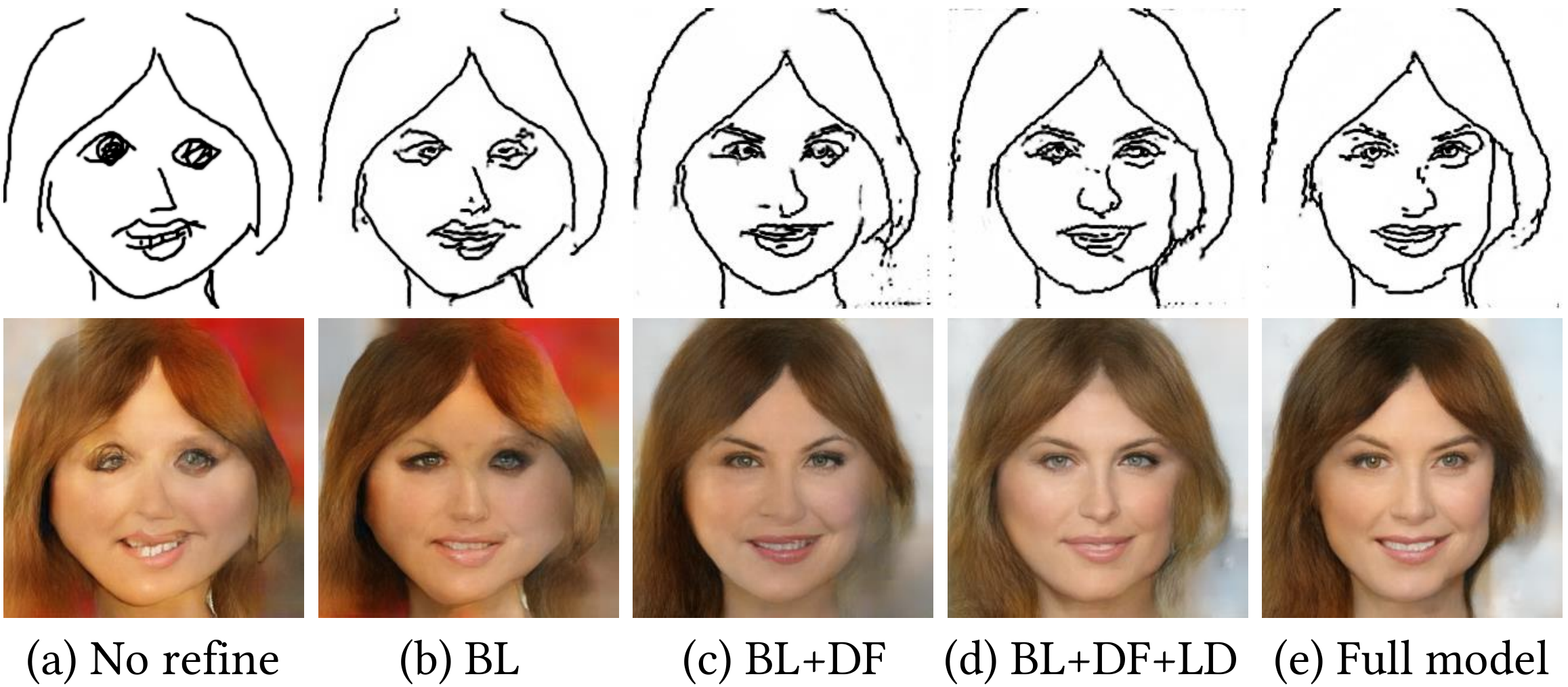}
  \caption{Effect of rough sketch models. (a) Input sketch and generated image without refinement. (b)-(e) Refinement results using different rough sketch models. (b) Baseline: edge dilation with a fixed single dilation radius. (c) Baseline + line deformation. (d) Baseline + line deformation and discarding. (e) Edge dilation with multiple radii + line deformation and discarding.}\label{fig:ablation-augment}
\end{figure}

\textbf{Rough sketch modelling}. We first examine the effect of our dilation-based sketch modelling, which is the key of our sketch refinement.
In Fig.~\ref{fig:ablation-augment}, we perform a comparison between different rough sketch models.
The dilation prompts the network to infer the facial details. Then the line deformation and discarding force the network to further infer and complete the accurate facial structures, respectively.
In Fig.~\ref{fig:ablation-augment}(e), we observe an improvement brought by learning multiple refinement levels in one model over single level per model.~The reason might be that coarse-level refinement can benefit from the learned more robust fine-level features.

\begin{figure}[t]
  \centering
  \includegraphics[width=0.98\linewidth]{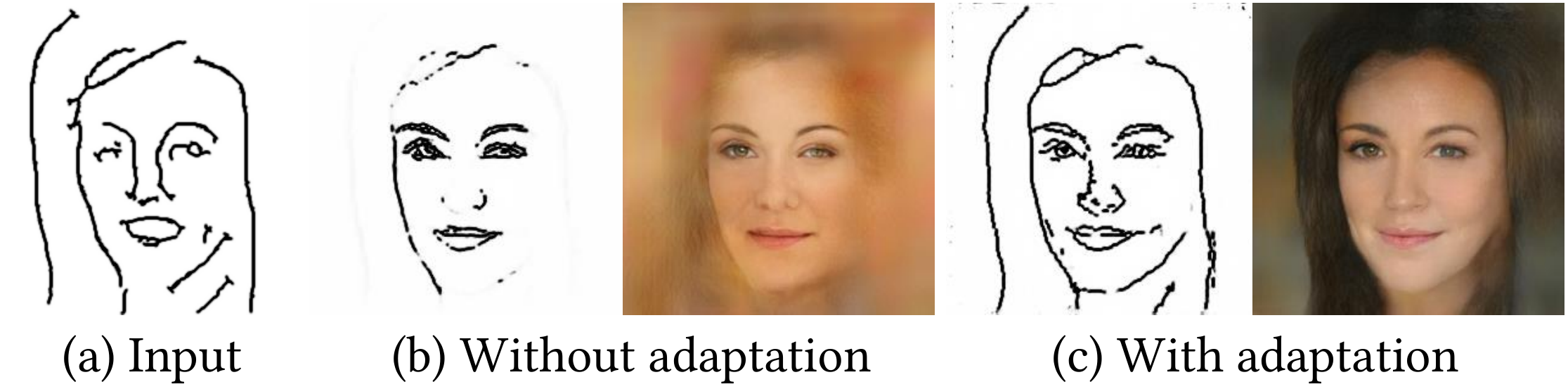}
  \caption{Effect of adaptation to $F$.}\vspace{-3mm}\label{fig:ablation-adaptation}
\end{figure}

\textbf{Adaptation to $F$}. Our generator $G$ is trained together with a fixed $F$, which adapts $G$ to $F$ to improve the quality of the ultimate output. To verify the effect of the adaptation, we train a model without the loss terms related to $I_{out}$ in Eqs.~(\ref{eq:rec_loss}) and  (\ref{eq:feat_loss}). Fig.~\ref{fig:ablation-adaptation} presents the comparison of our model with and without adaptation.
The sketch result without adaptation has its structure refined but some lines become indistinct. The reason might be the low proportion of the line region in the sketch.
Through adaptation, $G$ is motivated to generate sketches that are fully perceivable by $F$, which actually acts as a sketch-version perceptual loss~\cite{Johnson2016Perceptual}, resulting in distinct lines and high-quality photos.

\begin{figure}[t]
  \centering
  \includegraphics[width=0.98\linewidth]{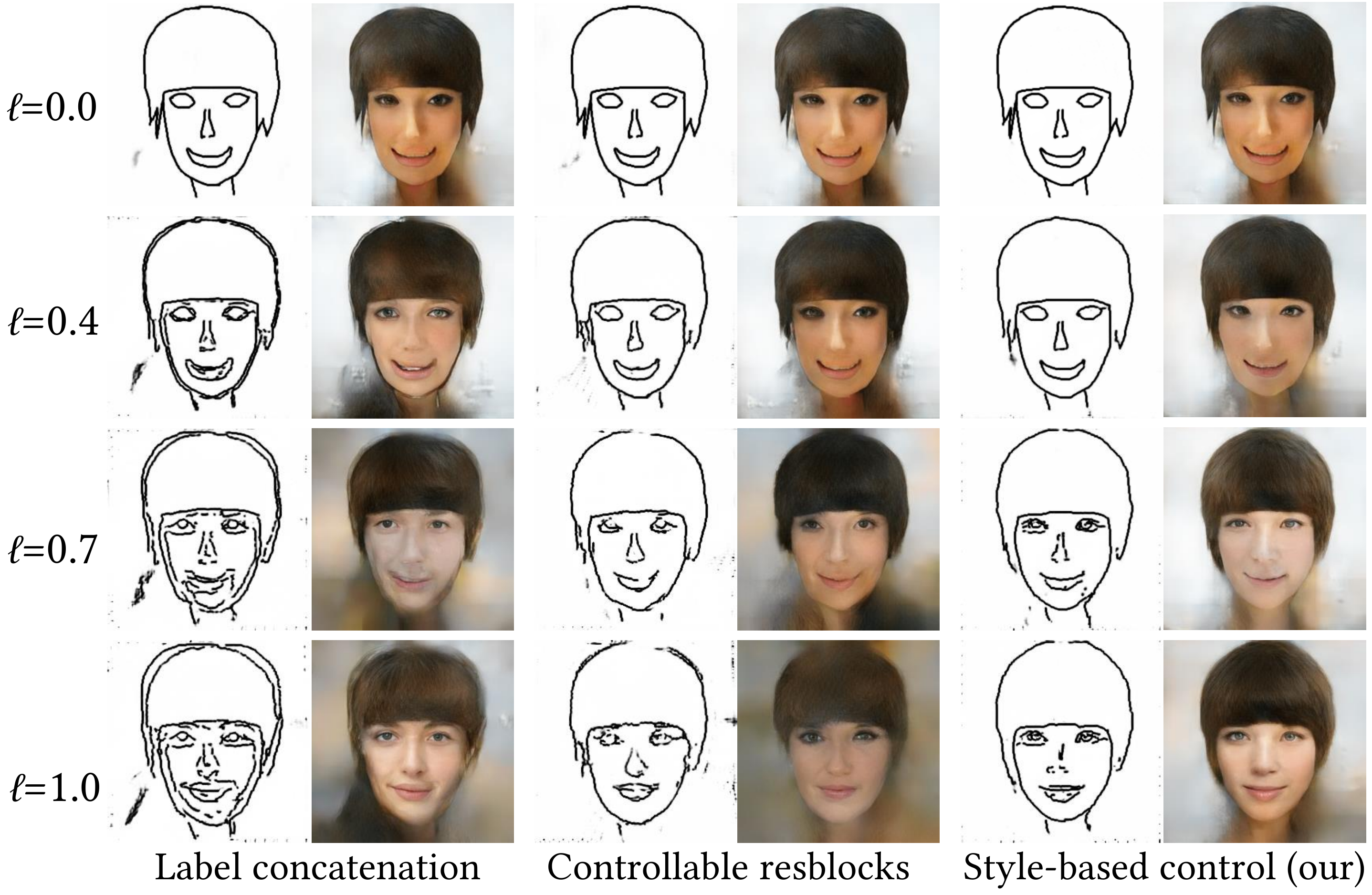}
  \caption{Visual comparison on label conditioning.}\label{fig:ablation-adaIN}
\end{figure}

\textbf{Refinement level control}. We compare the proposed style-based conditioning with label concatenation and controllable resblock~\cite{Yang2019Controllable} in Fig.~\ref{fig:ablation-adaIN}.
Label concatenation yields stacking lines like those in draft sketches.
By comparison, controllable resblock interpolates the resblock features at two extremes to achieve level control, which generates cleaner lines but still rough facial details.
Our style-based conditioning surpasses controllable resblock in adaptive channel-wise control,
which provides strongest results in both well-structured sketches and realistic photos.

\subsection{Applications}

\begin{figure}[t]
  \centering
  \includegraphics[width=0.98\linewidth]{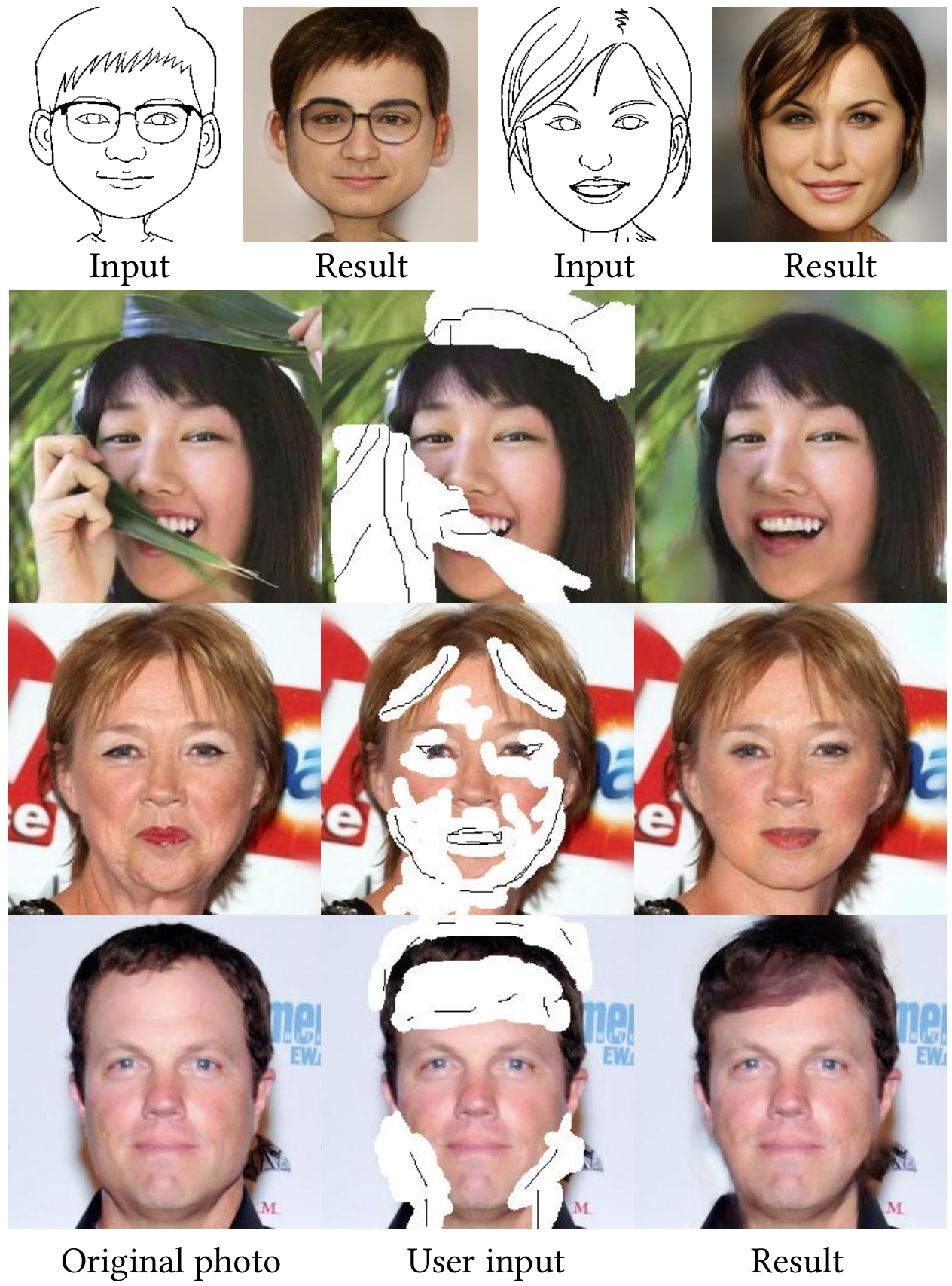}
  \caption{Applications. From top to bottom: cartoon to photo, object removal, rejuvenation and deep plastic surgery. }\label{fig:app}\vspace{-3mm}
\end{figure}

Fig.~\ref{fig:app} shows various results with sketch inputs.
Our model shows certain robustness on realistic photo rendering from cartoons. In the image editing setting, our model is inherently able to remove undesired objects, where the user guidance empowers the model to handle extremely complex scenes.
Finally, users can purposely perform ``plastic surgery'' digitally, such as removing wrinkles, lifting the eye corners. Alternatively, amateurs can intuitively edit the face with fairly coarse sketches to provide a general idea, such as face-lifting and bangs, and our model will tolerate the drawing errors and suggest a suitable ``surgery'' plan.

\subsection{Performance on Other Datasets}

\begin{figure}[t]
  \centering
  \includegraphics[width=0.98\linewidth]{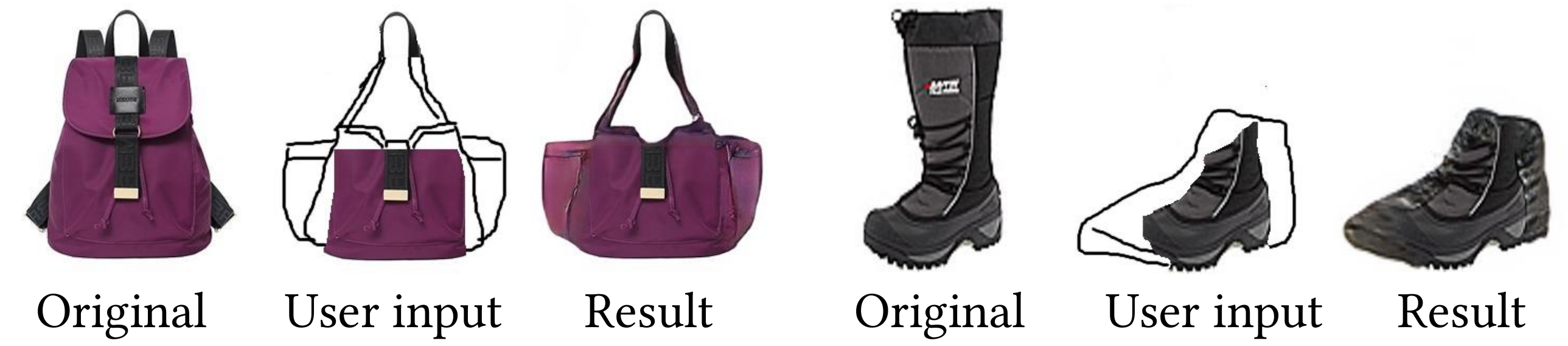}
  \caption{Performance on handbag and shoe datasets.}\label{fig:shoe}
\end{figure}

In addition to the facial images, we further present our results on the handbag and shoe datasets provided by~\cite{Isola2017Image}. 
The results are shown in Fig.~\ref{fig:shoe}, where our model can effectively modify the style of the handbags and the shoes.

\begin{figure}[t]
  \centering
  \includegraphics[width=0.98\linewidth]{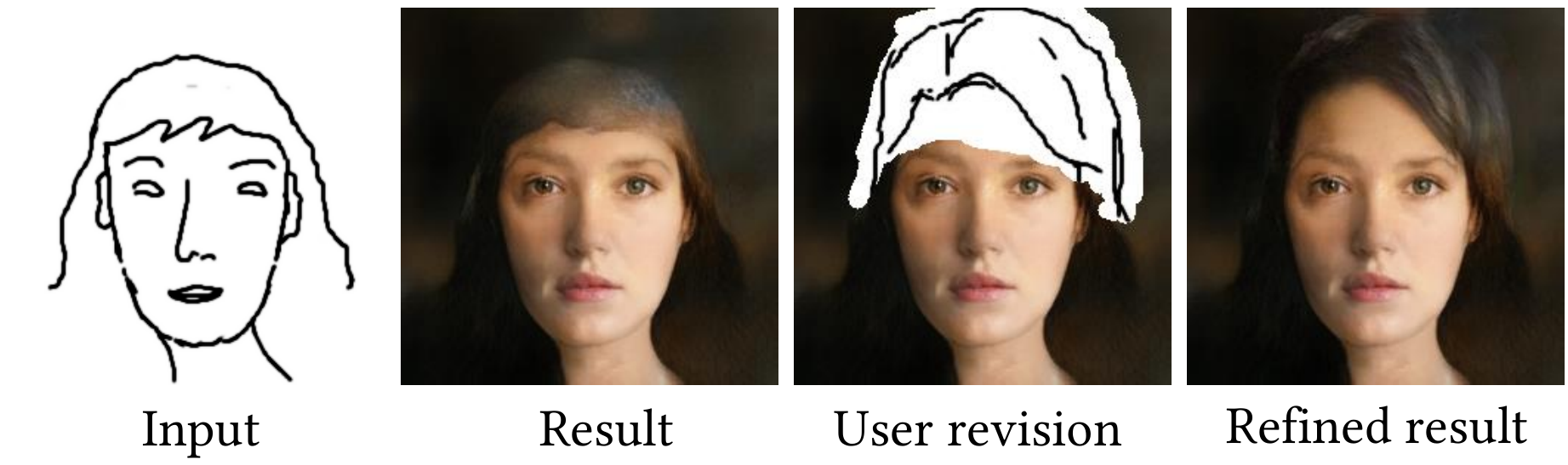}
  \caption{Failure case and user interactive revision to the position of the hairline and the top of the head.}\label{fig:fail}\vspace{-3mm}
\end{figure}


\section{Conclusion and Future Work}

In this paper, we raise a new  a  new  problem  of  controllable  sketch-based  image  editing,  to  adapt  edge-based models to human-drawn sketches, and present a novel dilation-based sketch refinement method.
Modelling the rough sketch as a drawable region via edge dilation, the network is effectively trained to infer accurate structural information.
Leveraging the idea of style transfer, our network is able to undo the edge dilation of different levels in a destylization manner for multi-level refinement control.
We validate by experiments the effectiveness and robustness of our method. Serving as a plug-in, our model can greatly improve the performance of edge-based models on the sketch inputs.

While our approach has generated appealing results, some limitations still exist.
First, our method cannot revise the structural error that exceeds the maximum allowable radius.
This problem can be possibly solved by user interaction, where users can modify the input sketch when the output is still unsatisfactory under the maximum refinement level as shown in Fig.~\ref{fig:fail}.
Second, when $\ell$ is large, the dilation operation will merge lines that are close to each other, which inevitably loses some structural details.
One solution is to use adaptive spatially varied dilation radii, which we would like to explore in future work.


{\small
\bibliographystyle{ieee_fullname}
\bibliography{egbib}
}

\clearpage

\onecolumn

\begin{minipage}{\textwidth}\centering
\Large{\textbf{Deep Plastic Surgery: Robust and Controllable Image Editing with\\ Human-Drawn Sketches}}\\
\vspace{10mm}
\large{\textbf{Supplementary Material}}\\
\end{minipage}

\vspace{10mm}
\begin{minipage}{1.0\textwidth}
As supplementary material of our paper, we present the following contents:
\vspace{4mm}
\begin{itemize}
  \item Detailed network architectures. (Fig.~\ref{fig:network})\vspace{1mm}
  \item Comparison with state-of-the-art methods. (Figs.~\ref{fig:comparison-IP-1}$-$\ref{fig:comparison-SN-2})\vspace{1mm}
  \item Quantitative evaluation (user study). (Table~\ref{tb:1})\vspace{1mm}
\end{itemize}
\end{minipage}
\clearpage

\section{Detailed network architectures}

\begin{figure}[H]
  \centering
  \includegraphics[width=0.86\linewidth]{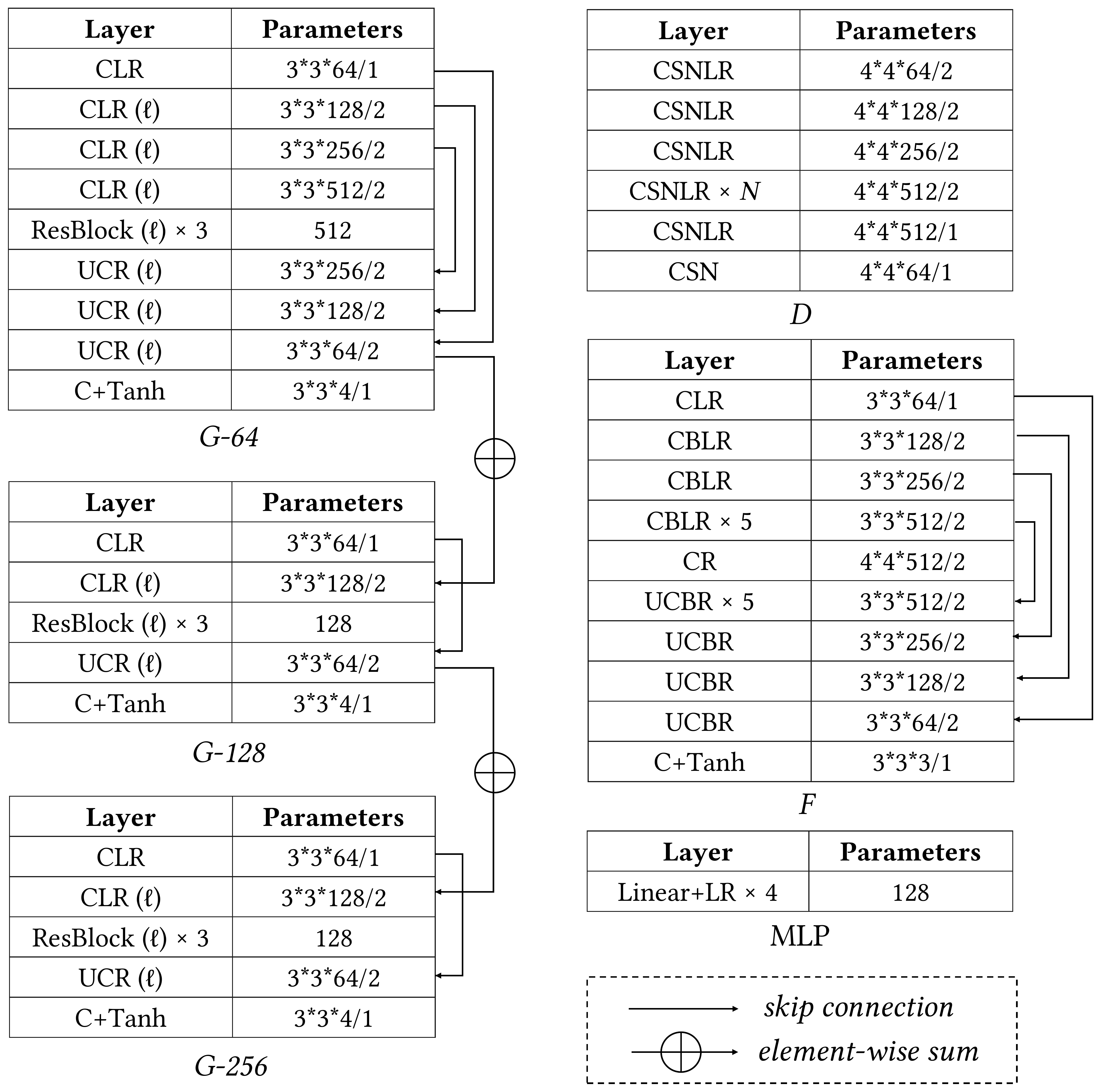}
  \caption{Overview of network architectures.}\label{fig:network}
\end{figure}

Our generator $G$ utilizes the fully convolutional Encoder-ResBlocks-Decoder architecture as in~\cite{Johnson2016Perceptual}. The discriminator $D$ follows the SN-PatchGAN~\cite{yu2018free} for stable and fast training.
Finally, we use pix2pix~\cite{Isola2017Image} as our edge-based baseline model $F$.
The detailed network architectures are shown in Fig.~\ref{fig:network}, where
``C'' denotes a Convolution layer,
``CSN'' denotes a Convolution-SpectralNorm layer~\cite{miyato2018spectral},
``CB'' denotes a Convolution-BatchNorm layer,
the prefix ``U'' denotes an Upsampling layer,
the suffix ``R'' and ``LR'' denote ReLU and LeakyReLU layers, respectively.
Finally, we use ``LayerName~($\ell$)'' to denote the convolutional layer is followed by the AdaIN layer for refinement level control.
We use ``$k*k*c/s$'' to indicate that the convolutional layer has $c$ filters with a spatial size of $k*k$ and stride $s$.
We follow pix2pixHD~\cite{Wang2017High} to use multi-scale $G$ and $D$ to gradually refine sketches from $64\times64$ to $128\times128$ to $256\times256$. Note that loss terms related to $I_{out}$ are computed only in $256\times256$ resolution since we usually only have $F$ for the target resolution available in real application.
For the image resolution of $64$, $128$ and $256$, the number $N$ of ``CSNLR'' of discriminator $D$ is set to $0$, $1$ and $2$, respectively. For each resolution, we first train our network with $\ell=1$ for $30$ epoches, and then train with uniformly sampled $\ell\in[0,1]$ for $200$ epoches. $F$ is trained with a discriminator whose architecture is based on $D$ with an additional linear layer to map the output tensor to a score.

\clearpage

\section{Comparisons with State-of-the-Art Methods}

\textbf{Face editing}.~Figs.~\ref{fig:comparison-IP-1}$-$\ref{fig:comparison-IP-2} present the qualitative comparison on face editing with two state-of-the-art inpainting models: DeepFillv2~\cite{yu2018free} and SC-FEGAN~\cite{jo2019sc}. The released DeepFillv2 uses no sketch guidance, which means the reliability of the input sketch is set to zero ($\ell=\infty$). Despite being one of the most advanced inpainting models, DeepFillv2 sometimes fails to repair the fine-scale facial structures well, indicating the necessity of user guidance. SC-FEGAN, on the other hand, totally follows the inaccurate sketch and yields weird faces, due to unrealistic details contained in the rough sketches. Our method yields more natural and realistic facial details.

\begin{figure}[H]
  \centering
  \includegraphics[width=0.98\linewidth]{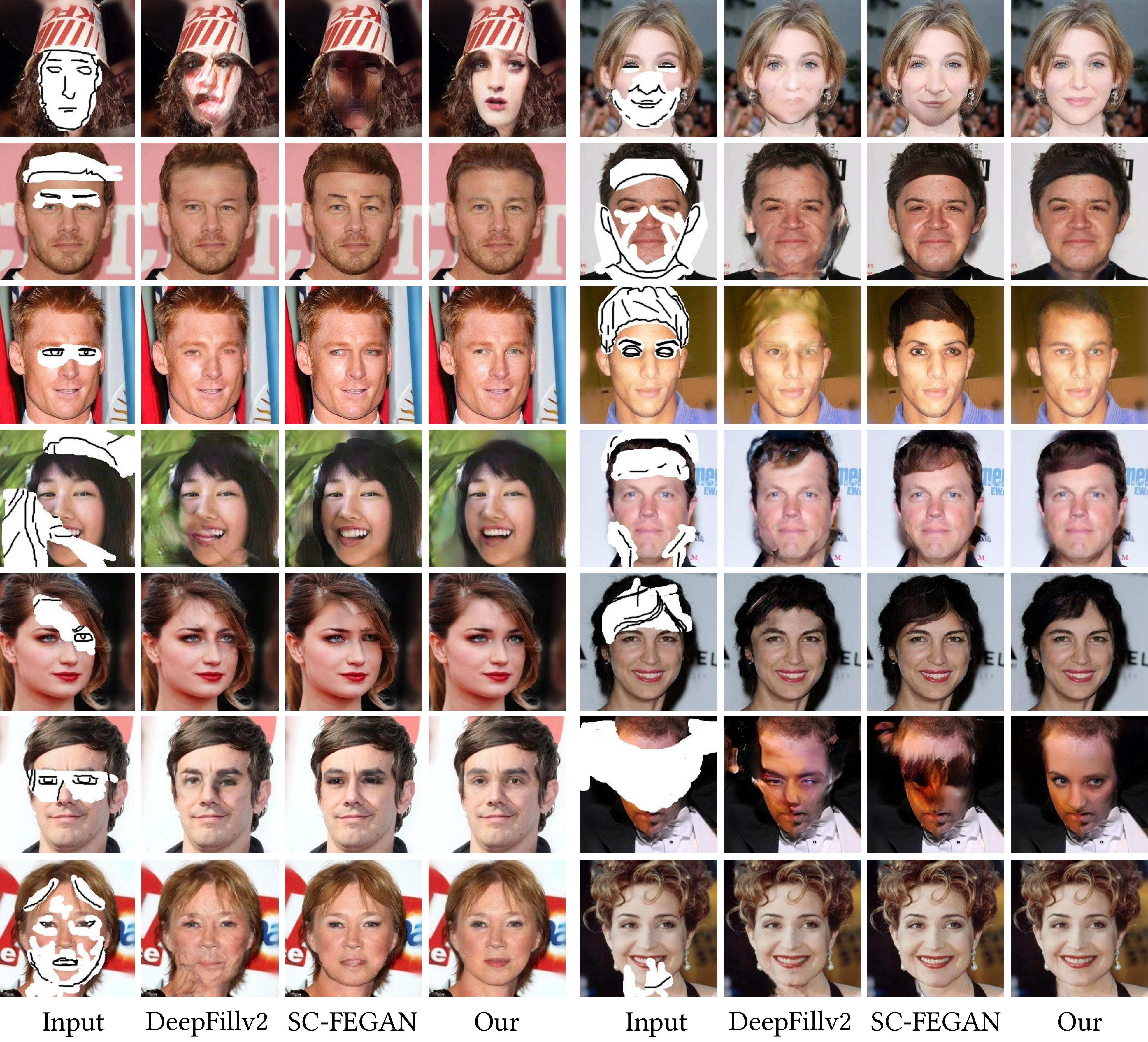}
  \caption{Comparison with state-of-the-art methods on face edting (Part I). For each group, from left to right: User inputs, results by DeepFillv2~\cite{yu2018free}, results by SC-FEGAN~\cite{jo2019sc} and our results.}\label{fig:comparison-IP-1}
\end{figure}

\clearpage

\begin{figure}[H]
  \centering
  \includegraphics[width=0.98\linewidth]{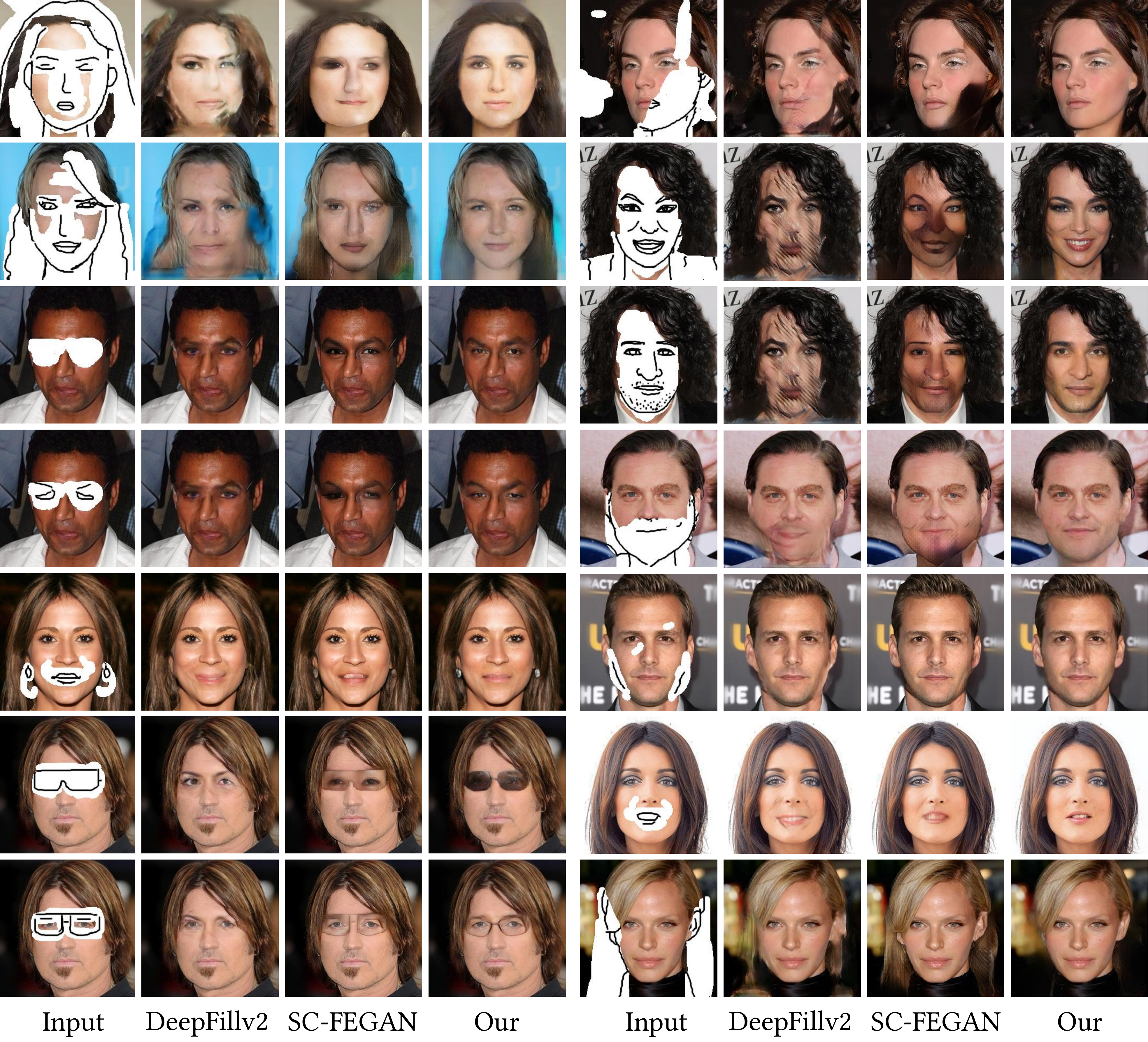}
  \caption{Comparison with state-of-the-art methods on face edting  (Part II). For each group, from left to right: User inputs, results by DeepFillv2~\cite{yu2018free}, results by SC-FEGAN~\cite{jo2019sc} and our results.}\label{fig:comparison-IP-2}
\end{figure}

\clearpage

\textbf{Face synthesis}. Figs.~\ref{fig:comparison-SN-1}$-$\ref{fig:comparison-SN-2} show the qualitative comparison on face synthesis with two state-of-the-art image-to-image translation models: BicycleGAN~\cite{zhu2017toward} and pix2pixHD~\cite{Wang2017High}. As expected, both models synthesize facial structures that strictly match the inaccurate sketch inputs, producing poor results. Our model takes sketches as ``useful yet flexible'' constraints, and strikes a good balance between authenticity and consistency with the user guidance.

\begin{figure}[H]
  \centering
  \includegraphics[width=0.98\linewidth]{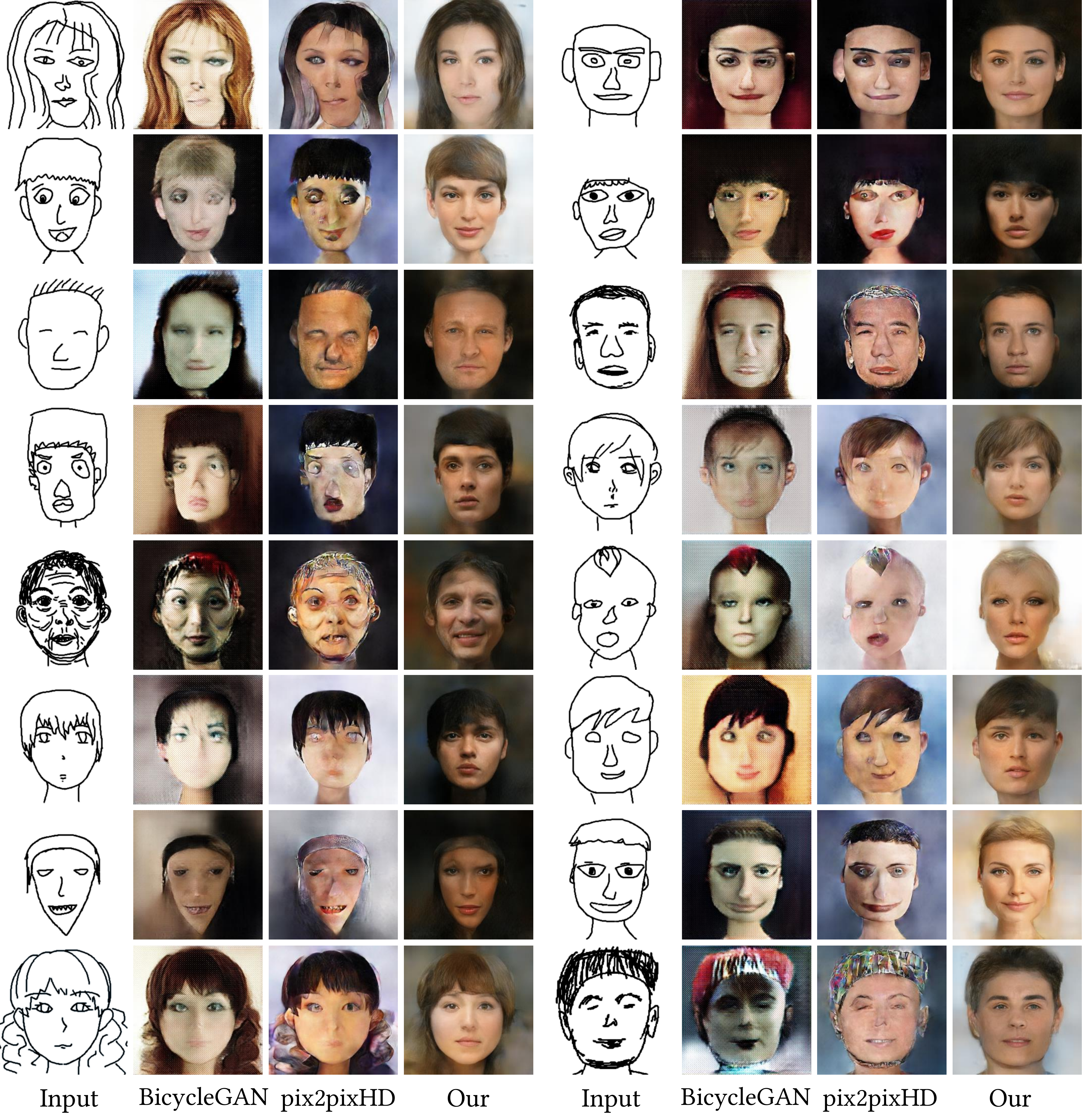}
  \caption{Comparison with state-of-the-art methods on face synthesis (Part I). For each group, from left to right: User inputs, results by BicycleGAN~\cite{zhu2017toward}, results by pix2pixHD~\cite{Wang2017High}, and our results.}\label{fig:comparison-SN-1}
\end{figure}

\clearpage

\begin{figure}[H]
  \centering
  \includegraphics[width=0.98\linewidth]{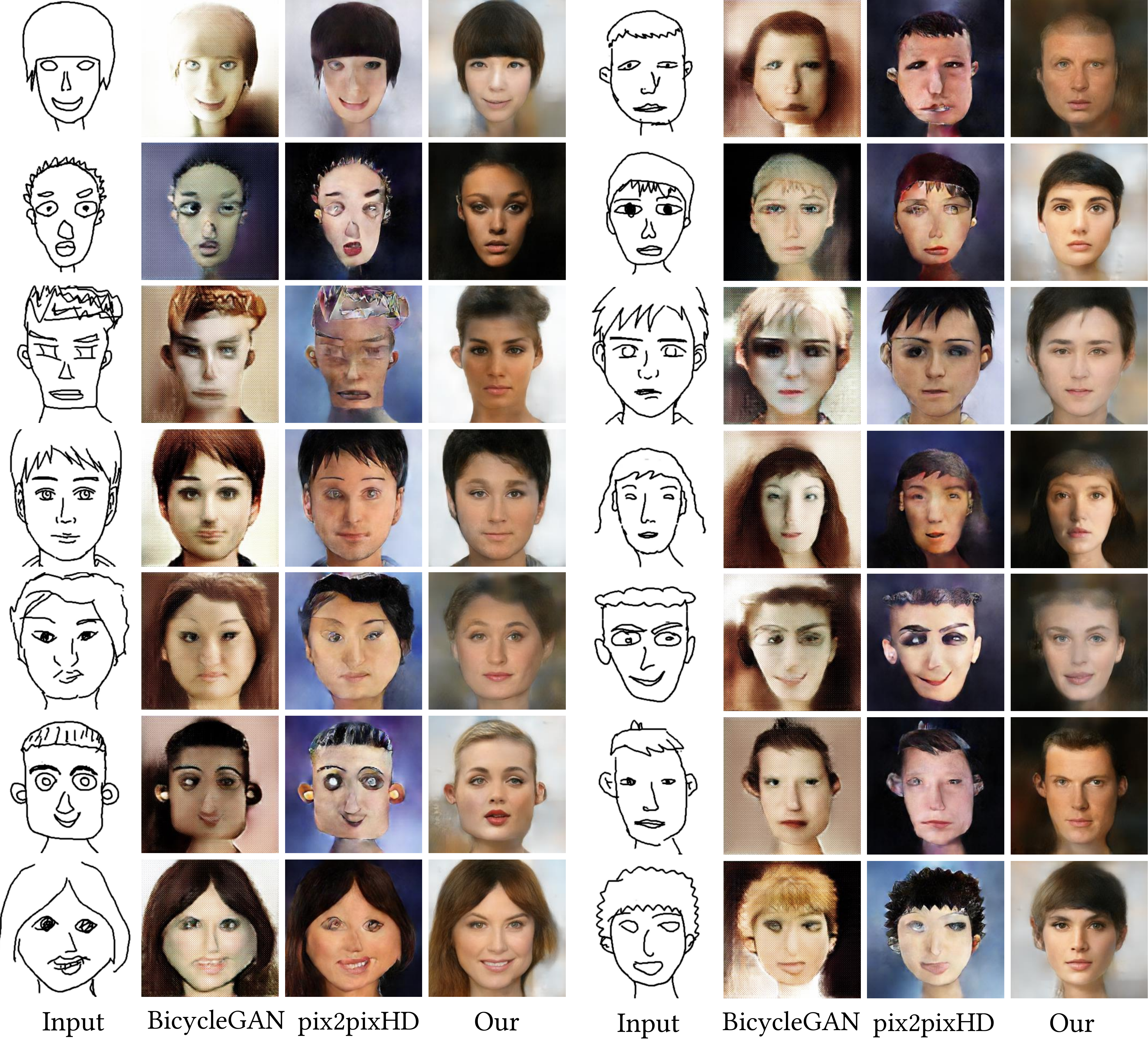}
  \caption{Comparison with state-of-the-art methods on face synthesis (Part II). For each group, from left to right: User inputs, results by BicycleGAN~\cite{zhu2017toward}, results by pix2pixHD~\cite{Wang2017High}, and our results.}\label{fig:comparison-SN-2}
\end{figure}

\clearpage

\section{Quantitative Evaluation}

To better understand the performance of the compared methods, we perform user studies for quantitative evaluations. Participants are shown the 28 face editing and 30 face synthesis cases in Figs.~\ref{fig:comparison-IP-1}$-$\ref{fig:comparison-SN-2}.
Each subject is asked to select one from the three results that best balances the sketch faithfulness with the output verisimilitude. To ensure the fairness, the orders of three methods randomly change every round. A total of 20 subjects participate in this study and a total of 1,160 selections are tallied.
The preference ratio is used as the evaluation metrics. It is calculated by:
\begin{equation}
  \text{preference ratio of Method}~A = \frac{~\text{The total number of times Method}~A~\text{was selected}~}{~\text{The total selection number}~}.
\end{equation}
According to the definition, if Method $A$ performs significantly better than all other methods, its mean preference ratio can reach $1.0$.
As shown in Table~\ref{tb:1}, for the task of face editing, the proposed method obtains the best average preference ratio of $0.730$, while the average scores of DeepFillv2~\cite{yu2018free} and SC-FEGAN~\cite{jo2019sc} are $0.032$ and $0.238$, respectively. For the task of face synthesis, the proposed method achieves preference ratios above $0.5$ in all cases, which means our method is steadily preferred by the users. The proposed method obtains the best average preference ratio of $0.945$, while the average scores of BicycleGAN~\cite{zhu2017toward} and pix2pixHD~\cite{Wang2017High} are $0.024$ and $0.031$, respectively. This user study quantitatively verifies the superiority of our method.

\begin{table} [H]
\caption{User preference ratio of state-of-the-art methods. The best score in each row is marked in bold.}\vspace{1mm}
\label{tb:1}
\centering
\begin{tabular}{c|ccc||c|ccc}
\toprule
\multicolumn{4}{c||}{Face Editing} & \multicolumn{3}{c}{Face Synthesis} \\
\midrule
ID & DeepFillv2 & SC-FEGAN & ~~~Ours~~~ & ID & BicycleGAN & pix2pixHD & ~~~Ours~~~ \\
\midrule
1 & 0.00  & 0.00  & \textbf{1.00}  & 1 & 0.15  & 0.10  & \textbf{0.75} \\
2 & 0.10  & \textbf{0.45}  & \textbf{0.45}  & 2 & 0.05  & 0.00  & \textbf{0.95} \\
3 & 0.20  & 0.25  & \textbf{0.55}  & 3 & 0.05  & 0.05  & \textbf{0.90} \\
4 & 0.00  & \textbf{0.50}  & \textbf{0.50}  & 4 & 0.00  & 0.00  & \textbf{1.00} \\
5 & 0.15  & 0.00  & \textbf{0.85}  & 5 & 0.00  & 0.25  & \textbf{0.75} \\
6 & 0.00  & 0.20  & \textbf{0.80}  & 6 & 0.00  & 0.10  & \textbf{0.90} \\
7 & 0.10  & \textbf{0.65}  & 0.25  & 7 & 0.05  & 0.00  & \textbf{0.95} \\
8 & 0.00  & \textbf{0.65}  & 0.35  & 8 & 0.00  & 0.00  & \textbf{1.00} \\
9 & 0.00  & 0.10  & \textbf{0.90}  & 9 & 0.00  & 0.00  & \textbf{1.00} \\
10 & 0.00  & \textbf{0.60}  & 0.40  & 10 & 0.00  & 0.00  & \textbf{1.00} \\
11 & 0.05  & \textbf{0.60}  & 0.35  & 11 & 0.00  & 0.10  & \textbf{0.90} \\
12 & 0.00  & 0.00  & \textbf{1.00}  & 12 & 0.00  & 0.00  & \textbf{1.00} \\
13 & 0.00  & \textbf{0.60}  & 0.40  & 13 & 0.00  & 0.15  & \textbf{0.85} \\
14 & 0.00  & \textbf{0.50}  & \textbf{0.50}  & 14 & 0.00  & 0.00  & \textbf{1.00} \\
15 & 0.00  & 0.00  & \textbf{1.00}  & 15 & 0.00  & 0.00  & \textbf{1.00} \\
16 & 0.00  & 0.00  & \textbf{1.00}  & 16 & 0.00  & 0.00  & \textbf{1.00} \\
17 & 0.00  & 0.15  & \textbf{0.85}  & 17 & 0.00  & 0.00  & \textbf{1.00} \\
18 & 0.00  & \textbf{0.50}  & \textbf{0.50}  & 18 & 0.05  & 0.00  & \textbf{0.95} \\
19 & 0.00  & 0.00  & \textbf{1.00}  & 19 & 0.05  & 0.00  & \textbf{0.95} \\
20 & 0.05  & 0.10  & \textbf{0.85}  & 20 & 0.00  & 0.00  & \textbf{1.00} \\
21 & 0.00  & 0.00  & \textbf{1.00}  & 21 & 0.05  & 0.00  & \textbf{0.95} \\
22 & 0.05  & 0.45  & \textbf{0.50}  & 22 & 0.00  & 0.00  & \textbf{1.00} \\
23 & 0.15  & 0.00  & \textbf{0.85}  & 23 & 0.10  & 0.00  & \textbf{0.90} \\
24 & 0.00  & 0.10  & \textbf{0.90}  & 24 & 0.05  & 0.00  & \textbf{0.95} \\
25 & 0.00  & 0.00  & \textbf{1.00}  & 25 & 0.00  & 0.10  & \textbf{0.90} \\
26 & 0.00  & 0.05  & \textbf{0.95}  & 26 & 0.05  & 0.05  & \textbf{0.90} \\
27 & 0.05  & 0.20  & \textbf{0.75}  & 27 & 0.05  & 0.00  & \textbf{0.95} \\
28 & 0.00  & 0.00  & \textbf{1.00}  & 28 & 0.00  & 0.00  & \textbf{1.00} \\
 &  &  &  & 29 & 0.00  & 0.00  & \textbf{1.00} \\
 &  &  &  & 30 & 0.00  & 0.10  & \textbf{0.90} \\
\midrule
Average & 0.032  & 0.238  & \textbf{0.730}  & Average & 0.024  & 0.031  & \textbf{0.945} \\
\bottomrule
\end{tabular}
\end{table}

\end{document}